\documentclass[sigconf]{acmart}

\usepackage{multirow}
\usepackage{booktabs}
\usepackage{adjustbox}
\usepackage[justification=justified,skip=3pt]{caption}

\AtBeginDocument{%
  \providecommand\BibTeX{{%
    \normalfont B\kern-0.5em{\scshape i\kern-0.25em b}\kern-0.8em\TeX}}}

\copyrightyear{2020}
\acmYear{2020}
\setcopyright{acmcopyright}\acmConference[CIKM '20]{Proceedings of the 29th ACM International Conference on Information and Knowledge Management}{October 19--23, 2020}{Virtual Event, Ireland}
\acmBooktitle{Proceedings of the 29th ACM International Conference on Information and Knowledge Management (CIKM '20), October 19--23, 2020, Virtual Event, Ireland}
\acmPrice{15.00}
\acmDOI{10.1145/3340531.3411918}
\acmISBN{978-1-4503-6859-9/20/10}

\begin{document}
\fancyhead{}

\title{Ranking Enhanced Dialogue Generation}
\author{Changying Hao${}^{1,2}$, Liang Pang${}^{1,2}$*, Yanyan Lan${}^{1,2}$, Fei Sun${}^{3}$, Jiafeng Guo${}^{1,2}$, Xueqi Cheng${}^{1,2}$}
\affiliation{%
   \institution{${}^{1}$CAS Key Lab of Network Data Science and Technology, \\ Institute of Computing Technology, Chinese Academy of Sciences, Beijing, China}
}
\affiliation{%
 \institution{${}^{2}$University of Chinese Academy of Sciences, Beijing, China}
}
\affiliation{%
 \institution{${}^{3}$Alibaba Group, Beijing, China}
}
\email{{haochangying18s, pangliang, lanyanyan, guojiafeng,  cxq}@ict.ac.cn, ofey.sf@alibaba-inc.com}




%

\begin{abstract}
How to effectively utilize the dialogue history is a crucial problem in multi-turn dialogue generation. Previous works usually employ various neural network architectures (e.g., recurrent neural networks, attention mechanisms, and hierarchical structures) to model the history. However, a recent empirical study by Sankar et al. has shown that these architectures lack the ability of understanding and modeling the dynamics of the dialogue history. For example, the widely used architectures are insensitive to perturbations of the dialogue history, such as words shuffling, utterances missing, and utterances reordering. To tackle this problem, we propose a Ranking Enhanced Dialogue generation framework in this paper. Despite the traditional representation encoder and response generation modules, an additional ranking module is introduced to model the ranking relation between the former utterance and consecutive utterances. Specifically, the former utterance and consecutive utterances are treated as query and corresponding documents, and both local and global ranking losses are designed in the learning process. In this way, the dynamics in the dialogue history can be explicitly captured. To evaluate our proposed models, we conduct extensive experiments on three public datasets, i.e., bAbI, PersonaChat, and JDC. Experimental results show that our models produce better responses in terms of both quantitative measures and human judgments, as compared with the state-of-the-art dialogue generation models. Furthermore, we give some detailed experimental analysis to show where and how the improvements come from. \let\thefootnote\relax\footnotetext{*Corresponding Author}
\end{abstract}






\maketitle

{\fontsize{8pt}{8pt} \selectfont
\textbf{ACM Reference Format:}\\
Changying Hao, Liang Pang, Yanyan Lan, Fei Sun, Jiafeng Guo, Xueqi Cheng. 2020. Ranking Enhanced Dialogue Generation. In {\it Proceedings of the 29th ACM International Conference on Information and Knowledge Management (CIKM '20), October 19--23, 2020, Virtual Event, Ireland.} ACM, New York, NY, USA, 10 pages. https://doi.org/10.1145/3340531.3411918}

\section{Introduction}

With the development of deep learning techniques and the acquisition of a massive amount of chat data, multi-turn dialogue generation systems have achieved great progress, and many technology companies have launched their chat-bots, such as Microsoft Xiaoice, Cortana, and Apple Siri. 
As the dialogue goes on, the number of its turns increases, accompanying with the emotion changing ~\cite{wei2019emotion,peng2019topic}, topic drifting ~\cite{kim2016exploring,xing2017topic}, and task switching ~\cite{wen2016network,eric2017key}. Therefore, the dialogue history embodies the high-level dynamics, and how to effectively use the dialogue history becomes crucial yet a big challenge in the task of multi-turn dialogue system. 

Most recent neural dialogue generation methods are based on the Seq2Seq~\cite{sutskever2014sequence} framework, which can be mainly categorized into two groups. The first group treats the whole history as a sequence of text~\cite{vinyals2015neural}. The historical utterances are directly concatenated in chronological order before the encoding process to generate a dialogue history representation. Typically, recurrent neural network~\cite{mikolov2010recurrent} and Transformer neural network~\cite{vaswani2017attention} are used to aggregate all the dialogue history to a single vector, and the outputs of the encoder are treated as the input representation for response generation.
The second group treats the dialogue history hierarchically~\cite{serban2016building, serban2017hierarchical, tian2017make, xing2018hierarchical, zhang2019recosa}. Each history utterance is encoded separately, and then another encoder is applied on these encoded utterances sequentially to generate the historical representation. Both word-level encoder and utterance-level encoder are modeled by either recurrent neural network or Transformer neural network, hence both words in utterance and utterances in the dialogue history are captured. However, \citeauthor{sankar2019neural} point out that neither the recurrent neural network nor Transformer neural network in the Seq2Seq framework can fully capture the dynamics. As previous models focus more on the generative loss, the dynamics are often modeled simply and implicitly by the recurrent structure or the positional embeddings, which is insufficient for thoroughly using the useful information implicated in dynamics.

To further enhance their understanding to the history, we propose to explicitly model the dynamics for multi-turn dialogue generation. Specifically, in multi-turn dialogue, dynamics can be represented as the flow of the semantic information in history utterances, for example, task steps in task-oriented dialogue or topic drifting in chit-chat. In most of the time, the history utterance follows the consistency attribute, that the next utterance can be predicted by the former utterances. As an initial step of modeling dynamics in dialogue, we ignore the special cases, while assuming that if the next utterance can be selected from candidates using the former utterances, the dynamic flow will be captured by the generation model. Thus, we introduce a \textbf{R}anking \textbf{E}nhanced \textbf{D}ialogue generation framework (RED) to explicitly model the internal utterance order in the dialogue history.

The RED framework mainly consists of three modules. Despite the traditional utterance representation module and response generation module, an additional ranking enhanced module is introduced to explicitly model the dynamics of the dialogue histories. 
Referring to the definition of ranking task in information retrieval, for each utterance in dialogue history, the former utterances can be treated as a query, in the meanwhile, the consecutive utterance right behind the query can be treated as the most relevant document. The suitability of the dynamic flow between them can be treated as relevance in information retrieval. In other words, the utterance ranking task aims to identify the next utterance from the candidates, given the former utterances as a query, where the candidates can be selected from the utterances after the query.
So the target of the ranking enhanced module is to evaluate the relevance between the query and documents, by making use of the ranking technique which has been well-explored in information retrieval~\cite{liu2009learning}. We design both local ranking loss and global ranking loss to capture such dynamics, by using local or global information for utterance ranking. Our proposed RED framework is general and can be easily incorporated with most of the recent dialogue generation models. 

To demonstrate the effectiveness of our proposed RED framework, we conduct experiments on three public datasets to compare with two types of state-of-the-art neural dialogue generation models. Experimental results show that the proposed models outperform existing ones, by incorporating the ranking enhanced module. The detailed analysis further proves that explicitly learning the dynamics of the dialogue history does help response generation. Moreover, it gives some explanations on where and how the improvements come from.


\section{Related Works}
In this section, we first describe the general form of the multi-turn dialogue generation task. Then, two categories of recent models that make use of dialogue history are introduced. 

\subsection{Multi-turn Dialogue Generation Task}
The multi-turn dialogue generation task provides a context that contains the history utterances of a conversation, while recent single turn dialogue only provides the nearest one, called the post. 
Therefore, the multi-turn dialogue generation task naturally satisfies a two-level hierarchy: a sequence of sub-sequences, and sub-sequences of tokens. In particular, a dialogue is modeled as a sequence of utterances (subsequences), with each utterance being a sequence of words.
Formally, suppose that we are given a collection of dialog history and its corresponding response pairs, where the collection $\mathcal{D}=\{(H_i, r_i)\}_{i=1}^N$ contains $N$ history-response pair instances. Each dialogue history $H_i$ contains $M$ utterances, $H_i=[h_1, h_2, \dots, h_M]$ and each utterance $h_j \in H_i$ composites by a sequence of words with length $K$, denotes as $h_j=[w_1, w_2, \dots, w_K]$. The $r_i=[v_1, v_2, \dots, v_L]$ denotes the response sentence of the $i$-th instance.
In a Seq2Seq framework, neural generation models are trained on the given collection, which parameterizes a probability distribution $P$, governed by parameters $\theta$, over the set of all possible dialogues of arbitrary lengths. The probability of generating the response $r$ can be defined as:
\begin{equation} \label{eq:seq_gen}
	P_\theta(r)=\prod_{l=1}^{L} P_\theta(v_l | v_{<l}, H),
\end{equation}
where $v_{<l}=[v_1, v_2, \dots, v_{l-1}]$ denotes the words before $v_l$ in the response. The task is analogous to language modeling, with the critical difference that dialogue generation is conditioned on history $H$. Sampling from the model can be performed as in standard language modeling: sampling one word at a time from the conditional distribution $P_\theta(v_l | v_{<l}, H)$ given the previously sampled words.

\subsection{Multi-turn Dialogue Generation Models}
Using the dialogue history effectively is vital for generating coherent and meaningful responses. Existing studies can be mainly separated into two categories, sequential way and hierarchical way depending on the different network structures for encoding the dialogue history information. 

In a sequential way, it treats all dialogue history as a whole sequence of text, where the words in each utterance are directly concatenated in its chronological order. This category merely applies single turn dialogue generation models in the multi-turn settings, just treating current dialogue history as a longer post.
\citeauthor{vinyals2015neural}~\cite{vinyals2015neural} first applies the recurrent neural network to the dialogue generation task, which considers all previous utterances equivalently. 
To summarize all the past information up to the last token representation suffers from the vanishing gradient effect. Therefore the short-term goals will dominate the output distribution~\cite{bengio1994learning}. In particular, for sequences with high variability, the models are likely to favor short-term predictions as opposed to long-term predictions.
\citeauthor{bahdanau2014neural}~\cite{bahdanau2014neural} introduces the attention mechanism into a recurrent neural network to tackle this problem. 
For simplification, a pure attention-based approach is proposed, namely Transformer~\cite{vaswani2017attention}. 
By eschewing recurrence and instead relying entirely on attention mechanism, Transformer draws global dependencies between input and output which makes it more parallelizable than RNNs. 
However, all of the sequential models ignore the two-level hierarchy structure of the multi-turn dialogue generation task, which provides rich information during the encoding part.

In a hierarchical way, \citeauthor{serban2016building}~\cite{serban2016building} first propose the hierarchical recurrent encoder-decoder architecture (HRED), they use an utterance level RNN to get the embedding of each history utterance, and use an inter-utterance level RNN to get the final embedding of the whole history by feeding the embedding of each history utterance as the inputs.
Considering that treating all history turns indiscriminately is not proper as the response is only usually related to some certain previous history turns, \citeauthor{tian2017make}~\cite{tian2017make} proposed a weighted sequence (WSeq) attention model for HRED, using the cosine similarity to measure the degree of the relevance between the utterance of the current turn and each of utterance of the previous history turns.
\citeauthor{xing2018hierarchical}~\cite{xing2018hierarchical} extend the HRED with the attention mechanism~\cite{bahdanau2014neural} to attend to important parts within and among utterances with word-level attention and utterance-level attention. \citeauthor{zhang2019recosa}~\cite{zhang2019recosa} propose a Transformer based model named ReCoSa to HRED, they use history turns self-attention to strengthen the encoder and use history-response attention to extract and use the relevant history turns for response generation.

However, the aforementioned works solely concentrate on exploring new network structures for modeling dialogue history, expecting the dynamic flow can be captured by well-designed network structures implicitly. 

\begin{figure}
    \centering
    \includegraphics[width=0.85\linewidth]{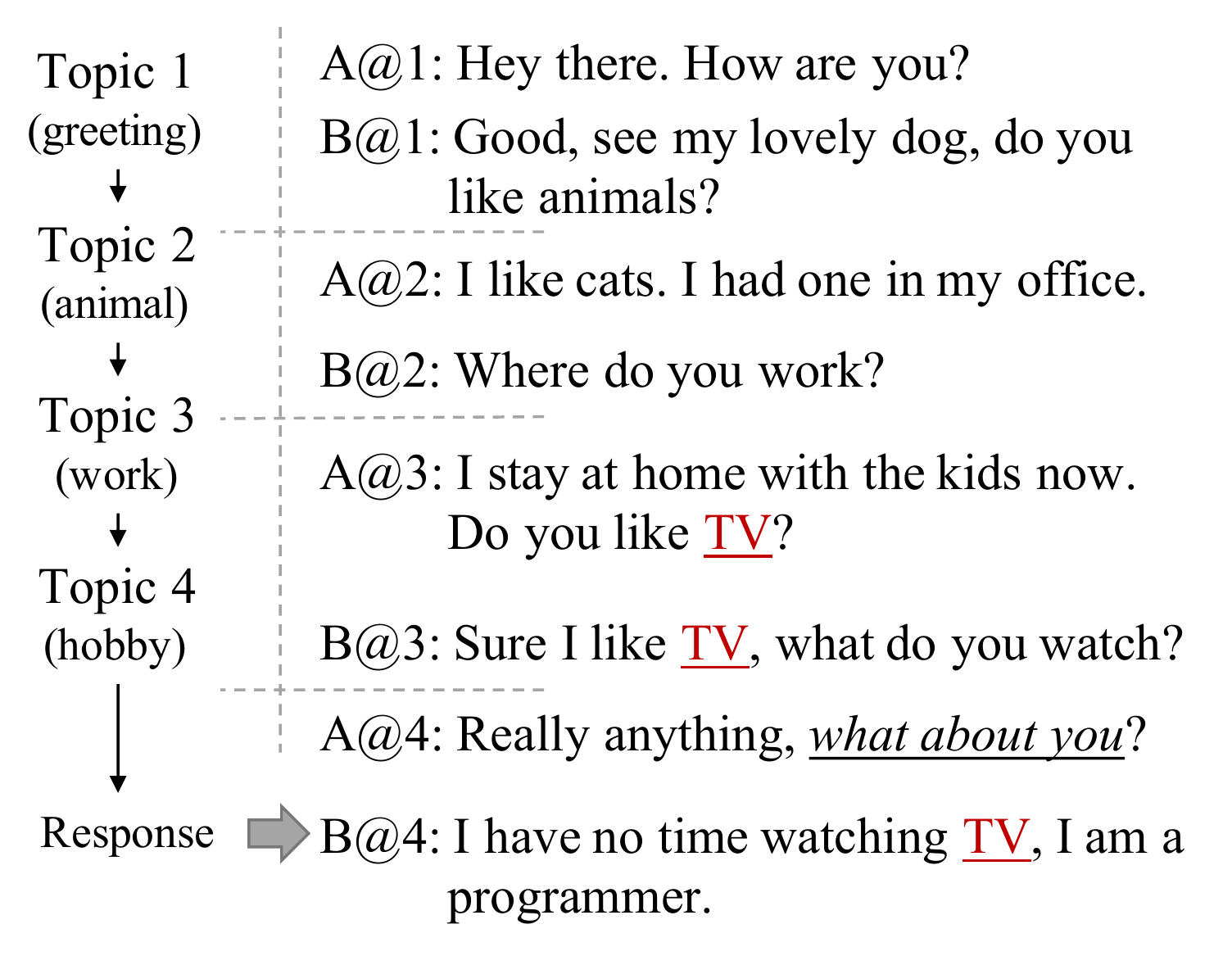}
    \caption{An example of dynamics in multi-turn dialogue.}
    \label{Fig:example}
\end{figure}

\section{Motivation}

Previous studies~\cite{li2017dailydialog, lowe2015ubuntu} take their effort to construct multi-turn dialogue datasets, expecting that richer contextual information can facilitate selecting and generating the next utterance~\cite{zhou2018multi}.
It is inspired by the behavior that human-generated responses are heavily dependent on the previous dialogue segments at different granularities (words, phrases, sentences, etc), both semantically and functionally, over multiple turns rather than one turn~\cite{lee2006complexity, traum1996utterance}. 
As we can see in Figure~\ref{Fig:example}, the post (A@4) is a general question `what about you?', thus dialogue history, such as `TV' entity plays an important role in generating the response.
In a typical multi-turn dialogue, there are several kinds of dialogue flows in the history dialogue to sustain the conversation, 1) the task-oriented conversation has a clear purpose which required to fulfill specific steps in order, 2) speaker can change from information-provider to information-seeker in turns, for example asking follow-up questions (B@1), 3) speakers often have different views about a topic (A@3), and by exchanging different proposals, they persuade and influence the other.
The above-described dialogue flows embed the dynamics inside, such as topic drifting in Figure~\ref{Fig:example}, from general to specific, from public to private.
Therefore, each utterance in history is predictable.

Recent works implicitly model the dynamics in history dialogue using the predefined network structures, such as recurrent neural network and transformer neural network. Recurrent neural network processes tokens step by step in history dialogue. Therefore, the output representation of the current token is a combination of previous token output representation and the current token input representation. Similarly, the Transformer structure utilizes positional embeddings to keep the order of utterance in mind. These model structures proposed for the sequential data assume to capture the dynamics implicitly.
Nevertheless, \citeauthor{sankar2019neural}~\cite{sankar2019neural} point out that neither the recurrent neural network nor the Transformer neural network in the Seq2Seq framework fully captures the dynamic flow in the dialogue history. In their study, authors make the perturbations to the dialogue history, but surprisingly find a marginal hurt to the generation performance. The increases to the perplexity of the generated response is too small, even though using drastic and unnatural modifications, including shuffling or reversing every utterance in the dialogue history. Such observations suggest that the well-known sequential model structures use far from all the information in the dialogue history that is available to them, not order information, let alone other complex dynamics.

Therefore, in this paper, we argue that the dynamics in history dialogue should be modeled explicitly. 
As we have seen in Figure~\ref{Fig:example}, the order of utterances reflects the simplest dynamic flow structure, any change of the dialogue history (A@1 - A@4) will destroy the dynamic flow in this conversation. In order to force utterances in order, predicting the next utterance using previous utterances is a good way which can be treated as a ranking problem. The previous utterance can be treated as a query and the next utterance can be treated as the most relevant document to the given query, while the other utterances are not as relevant as it. 
Thus we propose a ranking enhanced dialogue generation approach, in order to utilize dialogue history effectively.

\begin{figure*}
    \centering
    \includegraphics[width=0.7\linewidth]{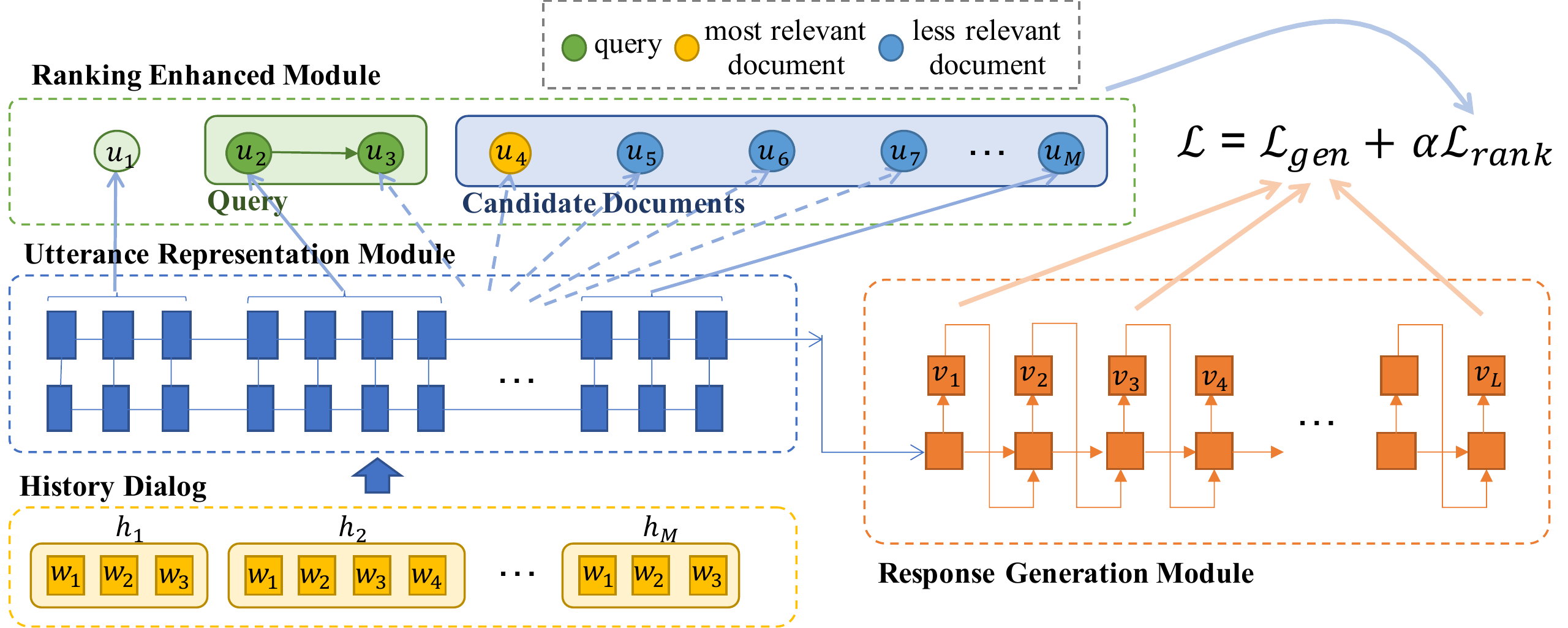}
    \caption{Overview architecture of RED model, RED with 2-order as an example.}
    \label{fig:framework}
\end{figure*}

\section{Our Approaches}
This section first introduces the architecture and major components involved in the proposed RED framework. As shown in Figure~\ref{fig:framework}, the RED framework mainly consists of three components: utterance representation module, response generation module, and ranking enhanced module. The former two modules are derived from traditional dialogue generation models, while the additional ranking enhanced module is the core part in RED distinguishing to them.

\textbf{Utterance representation module}: Given the dialogue history $H=[h_1, h_2, \dots, h_M]$, where $M$ is the number of history utterances. Each utterance is composited of words. We first get the word embedding for each word in all history utterances. For sequential models, the whole sequence of word embeddings are fed into an RNN encoder or a Transformer encoder according to the structure of different models. The encoder then outputs the representation for each word. We use the average of word representations for each history utterance as its representation. For hierarchical models, the word embeddings for each history utterance are first fed into an utterance level RNN encoder. The outputs of the utterance level RNN encoder are then fed into an inter-utterance level RNN encoder or Transformer encoder. We use the output of inter-utterance level encoder for each history utterance as its representation. The final representations of the dialogue history is denoted as $U=[u_1,u_2,\dots, u_M]$.

\textbf{Response generation module}: A typical decoding module in the Seq2Seq framework. In the training, our goal is to maximize the response probability defined in Equation~\ref{eq:seq_gen}. It takes the history dialogue and previously generated response words as the conditions, then maximizes the probability of choosing the next right response word. In the testing, one common way is to generate response word by word, and greedy pick the word with the highest probability at each time step.

\textbf{Ranking enhanced module}:
Directly constructed upon the utterance representation module, and shares the representations with the response generation module, as shown in the top of Figure~\ref{fig:framework}. The goal of ranking enhanced module is to make sure that the next utterance can be predicted by the utterances before, thereby maintaining the dynamic structure of dialogue history. To achieve this goal, we introduce a ranking perspective in this module, then construct the query, the most relevant document and less relevant documents using history utterances (the circles with different colors in Figure~\ref{fig:framework}). Therefore, the degree of learning dynamic flow can be evaluated by utilizing a list-wise ranking loss.

\subsection{Ranking Enhanced Module}
Assuming that simplest dynamic flow in dialogue history can be regarded as the chronological order between utterances, for the purpose of keeping all history utterances in order, we introduce a ranking perspective on the dialogue history.
Recalling that the top-1 relevant document ranking task in information retrieval defines a standard ranking framework. Given a query $q$ represents the user's information need and a set of $n$ documents $D=\{d_1, d_2, \dots, d_n\}$, where $d_r \in D$ represents the most relevant document. The goal of ranking is to maximize the probability $P(d_r|q)$.
Specifically, in the case of multi-turn dialogue generation, former utterances can be treated as query $q$, and the consecutive utterance right after the query can be treated as the most relevant document, while the others are less or not relevant. Note that the query is constructed by multiple former utterances (Section~\ref{section:qc}) and the relevant document is exactly the next utterance (Section~\ref{section:dc}).

\begin{figure}
    \centering
    \includegraphics[width=0.9\linewidth]{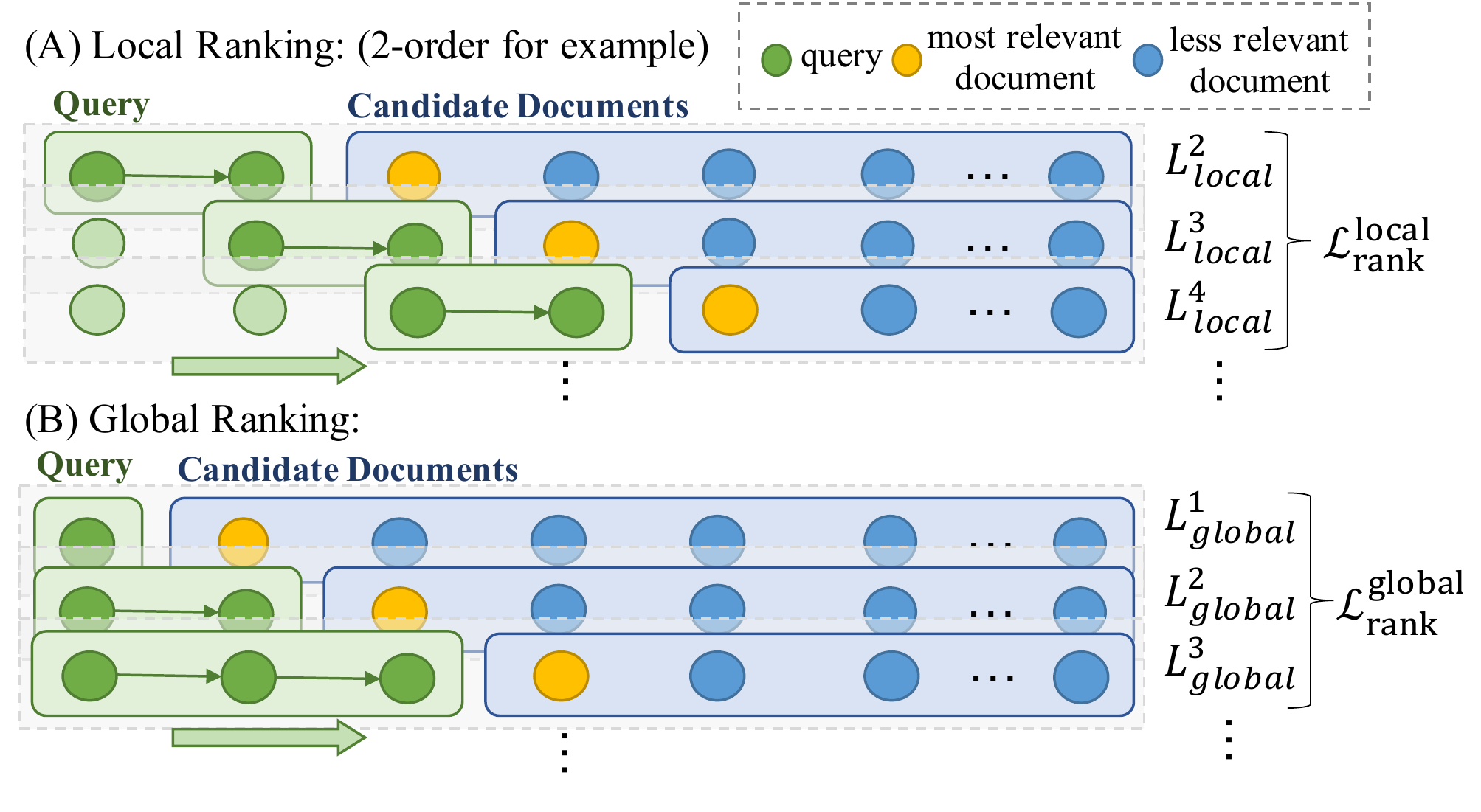}
    \caption{Two types of ranking enhanced module.}
    \label{fig:ranking}
\end{figure}

\subsubsection{Query Construction} \label{section:qc}

There are two approaches to construct the ranking query, i.e., local and global query construction. 
Within both ranking methods, a sequential structure LSTM is adopted to aggregate former utterances, in order to obtain a fixed-dimensional query representation. Despite the common structure, the scope of selecting former utterances is the main difference between the two ranking approaches.

\textbf{Query construction for local ranking}\label{section:local qc}
It is often the case that locally consecutive utterances belong to the same topic, or have a gradual and smooth topic drifting to keep coherence. To meticulously capture such local dynamic flow, we design the local query for ranking. Consider the utterance $u_i$ in a dialogue history, for constructing a $k$-order query in the local ranking method, the top $k-1$ former utterances closest to $u_i$ and the $u_i$ itself are used to construct the query. More precisely, the query representation $q_i$ is defined as,
\begin{equation}
    q_i=\mathrm{LSTM}(u_{i-k+1},\dots,u_{i}), i \in [k, M-1].
\end{equation}
where $u_i$ is the representation of the $i$-th utterance in dialogue history, $\mathrm{LSTM}$ is a function generating the last hidden state of the LSTM encoder.
As index $i$ varies from $k$ to $M-1$ in the history $H$, we can construct ${M-k}$ queries in total. The ${M-k}$ $k$-order queries can be obtained in a sliding window, where the window size is $k$. 
For example, as shown in Figure~\ref{fig:ranking} (A), the first query is constructed from utterances $[u_1, u_2]$, exactly fulfill the first window with size 2. When the window slides one step to the right, a new query $[u_2, u_3]$ is generated. When the window slides to the end, we get the last query $[u_{M-2}, u_{M-1}]$.

Note that in this work, $k$ is set to be 1, 2, 3, denote as 1-order query, 2-order query, and 3-order query, respectively. Note that when $k=1$, only $u_i$ itself is used to construct the 1-order query.

\textbf{Query construction for global ranking} 
To effectively utilize the previous dynamic flow till this history position and better modeling the global dynamic flow, different from using a fixed number of former utterances near the $i$-th utterance,
the global ranking method uses all former utterances before the $i$-th utterance and the $i$-th utterance itself to construct the query. 
We call this type of query as full-order query, the representations $q_i$ of which is then constructed as follows:
\begin{equation}
    q_i=\mathrm{LSTM}(u_{1},\dots,u_{i}), i \in [1, M-1].
\end{equation}

Similar to the local ranking query, $M-1$ queries can be constructed by using the different length of history, as in Figure~\ref{fig:ranking}~(B). 

\subsubsection{Documents Construction} \label{section:dc}
Given the dialogue history utterance representations $U=[u_1, u_2, \dots, u_M]$, for the $i$-th query $q_i$, the utterance $u_{i+1}$ is the most relevant document. 
Thus, the candidate document set contains all of the utterances after the $i$-th utterance, denoted as $U_{> i}=\{u_{i+1},\dots,u_{M}\}$. 
That is to say, each utterance in the candidate set can be seen as a unique document, and the query is used to rank the candidate documents according to their relevance with the query. The relevance here can be viewed as a certain type of manifestation for proper dynamic flow.
Specifically, $(q_i, u_{i+1})$ is the most relevant pair, while $\{(q_i, u_j),j>i+1\}$ are treated as less relevant or irrelevant pair.

\subsubsection{Ranking}\label{section:ranking}

For the local ranking method, the utterances in a sliding window described in~\ref{section:local qc} are used for constructing the query, all of the utterances after those utterances are candidate documents. For the local ranking methods using the k-order query, we can construct $M-k$ query-documents pairs. For the global ranking method, dialogue history can be simply divided into two consecutive parts, the former part is the query and the following part is the candidate documents, thus we construct $M-1$ query-document pairs.

Within each query-document pair, the Topk-ListMLE~\cite{xia2009top} is designed for correct ranking at the top-k positions in information retrieval. In our ranking task, given former utterances, we only need to rank the next utterance correctly at the top-1 position without taking the ranking position among utterances after the next utterance into account, because all of them are less relevant than the next utterance. Thus, we use Top1-ListMLE loss in the RED framework for our ranking task. 

Specially, in our ranking case, for the $i$-th query-documents pair, its query is  $q_i$, its candidate document set is $U_{> i}=\{u_{i+1},\dots,u_{M}\}$ and the ground-truth labels are $\{1,0,\dots,0\}$. We firstly get the ranking score for each element $u_t$ in the $U_{> i}$ by a ranking function $f_r$. The $f_r$ function is implemented in the following steps.
Firstly, we concatenate $q_i$ with each element in the $U_{>i}$ and get a sequence of concatenated representations,
$\bigl\{[q_i,u_{i+1}],[q_i,u_{i+2}],\dots,[q_i,u_{M-i}]\bigr\}$,
where $[\cdot, \cdot]$ represents the function that concatenates two vectors.

Secondly, for the utterance $u_t$ in $U_{>i}$, we use a fully connected layer to get its final ranking score. This layer takes the concatenated representation $[q_i,u_t]$ as the inputs and then outputs a real-value score for the utterance $u_t$. $f_r(u_t)$ can be formalized as,
\begin{equation}
f_r(u_t)=W^{r}\cdot [q_i,u_t]+b^{r},
\end{equation}
where $W^{r}$ and $b^{r}$ are the parameters of the fully connected layer.

After getting the ranking score for each utterance in the candidate documents, the Top1-ListMLE loss $L^i$ is defined as,
\begin{equation}
    L^i(U_{> i}; f_r) = -\log  \frac{\exp\bigl(f_r(u_{i+1})\bigr)}{\sum_{t=i+1}^{M}\exp\bigl(f_r(u_t)\bigr)} .
\end{equation}

For the ranking task, we design local ranking loss and global ranking loss for capturing the local and global dynamics implied in dialogue history turns. For the local ranking method, the local ranking loss for the $i$-th query-documents pair is denoted as $L_{\mathit{local}}^i$. For the global ranking method, the global ranking loss for the $i$-th query-documents pair is denoted as $L_{global}^i$.

Given a single history-response pair, the RED model whose query is k-order query designed for capturing local dynamics, the local ranking loss $\mathcal{L}_{\mathit{rank}}^{local}$ is defined as,
\begin{equation}
    \mathcal{L}_{\mathit{rank}}^{local} = \frac{1}{M-k}\sum_{i=k}^{M-1}L_{\mathit{local}}^i(U_{> i};f_r),
\end{equation}
and the RED model whose query is full\_order query designed for capturing global dynamics, the global ranking loss $\mathcal{L}_{\mathit{rank}}^{\mathit{global}}$ is defined as,
\begin{equation}
    \mathcal{L}_{\mathit{rank}}^{\mathit{global}} = \frac{1}{M-1}\sum_{i=1}^{M-1}L_{global}^i(U_{> i},f_r).
\end{equation}

\subsection{Objective Function and Training}

The RED framework appends ranking enhanced module to the original dialogue generation models, so the objective function contains two parts, including a traditional sequence generation loss, denotes as $\mathcal{L}_{gen}$ and a newly proposed ranking loss, denotes as $\mathcal{L}_{\mathit{rank}}$. The overall loss function is a linear combination of these two losses:
\begin{equation}
    \mathcal{L}=\mathcal{L}_{gen}+ \alpha \mathcal{L}_{\mathit{rank}},
\end{equation}
where $\alpha$ is a hyper-parameter that control the ratio of two losses. Moreover, the generative loss $\mathcal{L}_{gen}$ for response generation module is defined as,
\begin{equation}
    \mathcal{L}_{gen}=-\log(\prod_{l=1}^{L} P(v_l | v_{<l}, H))
\end{equation}
where $H$ is the given dialogue history and the target response is represented as $r = \{v_1, v_2, \dots, v_L \}$. 
The ranking loss $\mathcal{L}_{\mathit{rank}}$ can be set to either $\mathcal{L}_{\mathit{rank}}^{local}$ or $\mathcal{L}_{\mathit{rank}}^{\mathit{global}}$, to complete a local ranking task or a global ranking task, and can be seen as a regular type.
We jointly train the response generation module and the ranking enhanced module by sum up the generation loss and the ranking loss with a parameter $\alpha$, then we get the final loss function $\mathcal{L}$. 

The training for generation is a type of supervised training, and the training for ranking is indeed a type of self-supervised learning. We train the RED models with joint training loss using batched Stochastic Gradient Descent (SGD) with Adam ~\cite{kingma2014adam}. 

\section{Experiments}

Experiments are conducted on three public multi-turn dialogue datasets, i.e., bAbI, PersonaChat and JDC, including task-oriented datasets and open domain datasets in multiple languages, e.g English and Chinese, in order to thoroughly evaluate our proposed RED framework.
In this section, we first introduce experimental settings including dataset statistics, evaluation metrics and baselines description. Result evaluations, then, proposed in order to demonstrate the effectiveness and generalization of our RED framework. Furthermore, empirical analysis is conducted to show where and how the improvements come from.

\begin{table*}
\centering
\caption{Descriptive statistics of datasets, including the number of dialogues in train/valid/test sets, the total number of history-response pairs,the average number of turns, the average number of words per turn and vocabulary size}
\label{Table:datasets}
\begin{tabular}{lccccccc}
    \toprule
    Dataset & \#Train & \#Valid & \#Test &\#History-Response Pairs &\#Avg. Turns & \#Avg. Words & \# Vocabulary \\
    \midrule
    PersonaChat & 8939  & 1000  & 968  &162064 & 14.86 & 14.24 & 18745 \\
    JDC   & 99944 & 1549  & 1565 &980023 & 9.51  & 15.86 & 113503\\
    bAbI dialog & 1000  & 1000  & 1000  &110390 & 36.8  & 7.02  & 1109\\
    \bottomrule
\end{tabular}%
\end{table*}

\subsection{Experimental Settings}
\subsubsection{Datasets}
We conduct experiments on three multi-turn dialogue datasets with different styles, they are the bAbI dialog~\cite{bordes2016learning}, the PersonaChat ~\cite{zhang2018personalizing} and the Chinese customer service dataset (JDC)~\cite{zhang2019recosa} respectively. 
Each dataset is split into train/valid/test sets according to the previous works~\cite{zhang2019recosa,sankar2019neural}. Note that each multi-turn dialogue in the three datasets is processed to many history-response pairs with different history lengths. Table~\ref{Table:datasets} provides descriptive statistics about the three datasets.

\textbf{PersonaChat} is an open domain dataset with multi-turn chit-chat conversations between turkers who are each assigned a ``persona'' at random. It comprises 10.9k dialogs with an average of 14.86 turns per dialog. 

\textbf{JDC} The Chinese customer service dataset, named JDC, consists of 103058 dialogues published by the JD multi-turn dialogue challenge. This dataset is a collection of multi-turn dialogues between customer and customer service in the JD E-commerce platform.  The average turns per dialogue is 9.51.

\textbf{bAbI Dialog} is a synthetic goal-oriented multi-turn dialogue dataset consisting of 5 different tasks for restaurant booking with increasing levels of complexity. We consider Task 5 in our experiments since it is the hardest and is a union of all four tasks. The average turns per dialogue is 36.8.

\begin{table*}
\caption{Human evaluation on the PersonaChat dataset.}
\label{human evaluation}
\setlength{\tabcolsep}{0.9em}
\begin{tabular}{l|cc|cc|cc}
\toprule
& \multicolumn{2}{c|}{Informativeness} &  \multicolumn{2}{c|}{Fluency} & \multicolumn{2}{c}{Relevance} \\
\midrule
\multicolumn{1}{l|}{Models v.s. Transformer} & Win/Tie/Loss & Kappa & Win/Tie/Loss & Kappa & Win/Tie/Loss & Kappa \\
\midrule
- RED1\_Transformer & 50.13/44.73/5.14      &  0.493     &  23.02/56.31/20.67     &  0.578     &  25.47/54.04/20.49     &0.589 \\
- REDfull\_Transformer &   51.30/39.51/10.19    &  0.482     &  35.62/44.06/20.32   &  0.569    &   35.32/59.48/5.20    & 0.510\\
\midrule
\multicolumn{1}{l|}{Models v.s. ReCoSa} & Win/Tie/Loss & Kappa & Win/Tie/Loss & Kappa & Win/Tie/Loss & Kappa \\
\midrule
- RED1\_ReCoSa & 35.79/34.38/30.63  & 0.477 &   32.90/39.58/28.52    & 0.623      &   16.90/69.31/14.79    & 0.497 \\
- REDfull\_ReCoSa & 35.62/44.77/20.61      &  0.515     & 25.15/54.71/22.14      &   0.596    &   21.29/59.60/19.11    & 0.524 \\
\bottomrule
\end{tabular}%
\end{table*}

\begin{table*}
\caption{Examples of responses generated by baselines and RED models on the PersonaChat dataset.}
\label{case study}
\begin{tabular}{p{24em}|l}
\toprule
\multicolumn{1}{c|}{Multi-turn Dialogue History} & \multicolumn{1}{c}{Generated Responses} \\
\midrule
A: Hi . Just chilling to john denver after my work out .  & \textbf{Transformer:}               I am a cashier at the local department . \\
    \quad How about you ? & \textbf{RED1\_Transformer:}      I like music . I love rap music . \\
B: Hi , just studying for my college exam . Hows it going & \textbf{REDfull\_Transformer:}    I love country music . I love country music .  \\
\cmidrule{2-2}A: Good . What are you taking in college ? & \textbf{ReCoSa:}                      I am from georgia . I am in the navy . \\
B: What kind of music do you like to listen to ? & \textbf{RED1\_ReCoSa:}             I am a fan of 60s and 70s music . \\
A: John Denver . How about you ? & \textbf{REDfull\_ReCoSa:}          I like all kinds of music . I love all kinds of music . \\
\midrule
A: Scotch goes good with meat & \textbf{Transformer:}                I have a brother , he is a detective . \\
B: Scotch goes good with anything lol . & \textbf{RED1\_Transformer:}      I have a cat named after my daughter \\
A: I have a brother who likes scotch & \textbf{REDfull\_Transformer:}   I have a cat named gizmo . He is a big cat . \\
\cmidrule{2-2}B: I am an only child . & \textbf{ReCoSa:}                       I do not have any pets , I have a dog and a dog \\
A: I have just the one brother and 3 sisters & \textbf{RED1\_ReCoSa:}             I do not have any pets \\
B: Do you have any pets ? I am scared of dogs .  & \textbf{REDfull\_ReCoSa:}          I have a dog , I love dogs \\
\bottomrule
\end{tabular}%
\end{table*}

\begin{table}
    \caption{Performance of queries of different order for five models on the PersonaChat dataset. The best score is bolded.}
	\label{Table:exp_personachat}
	\begin{adjustbox}{max width=0.95\linewidth}
    \begin{tabular}{clccc}
    \toprule
    & Model & PPL & Dist-1 & Dist-2 \\
    \midrule
    \multirow{5}{*}{Seq2Seqs} 
            & Seq2Seq & 43.04 & 0.0035 & 0.0081 \\
    \cmidrule{2-5}      
            & RED1\_Seq2Seq & 42.90 & \textbf{0.0042} & \textbf{0.0110} \\    
            & RED2\_Seq2Seq & 43.01 & 0.0039 & 0.0099 \\
            & RED3\_Seq2Seq & \textbf{42.70} & 0.0040 & 0.0098 \\
            & REDfull\_Seq2Seq & 43.03  & 0.0034 & 0.0080 \\
    \midrule
    \multirow{5}{*}{SeqAtts} 
            & SeqAtt & 40.98 & 0.0111 & 0.0385 \\
    \cmidrule{2-5}      
            & RED1\_SeqAtt & 40.81 & \textbf{0.0115} & \textbf{0.0411} \\
            & RED2\_SeqAtt & 40.92 & 0.0102 & 0.0362 \\
            & RED3\_SeqAtt & \textbf{40.56} & 0.0107 & 0.0400 \\
            & REDfull\_SeqAtt & 40.96 & 0.0097 & 0.0334 \\
    \midrule
    \multirow{5}{*}{Transformers} 
            & Transformer & 40.48 & \textbf{0.0138} & \textbf{0.0559} \\
    \cmidrule{2-5}      
            & RED1\_Transformer & 39.49 & 0.0103 & 0.0352 \\
            & RED2\_Transformer & \textbf{39.45} & 0.0104 & 0.0380 \\
            & RED3\_Transformer & 39.54 & 0.0106 & 0.0401 \\
            & REDfull\_Transformer & 39.46 & 0.0121 & 0.0426 \\
    \midrule
    \multirow{5}{*}{HREDs} 
            & HRED & 44.96 & 0.0040 & 0.0105 \\
    \cmidrule{2-5}      
            & RED1\_HRED & 44.02 & \textbf{0.0044} & 0.0116 \\
            & RED2\_HRED &  \textbf{44.01}     &   0.0043    & \textbf{0.0123} \\
            & RED3\_HRED &    44.46   &   0.0036    & 0.0093 \\
            & REDfull\_HRED & 44.40      &   0.0033    & 0.0093 \\
    \midrule
    \multirow{5}{*}{ReCoSas} 
            & ReCoSa & 38.27  & 0.0066 & 0.0174\\
    \cmidrule{2-5}      
            & RED1\_ReCoSa & 37.89 & 0.0072 & 0.0204 \\
            & RED2\_ReCoSa & \textbf{37.57} & \textbf{0.0095} & \textbf{0.0281} \\
            & RED3\_ReCoSa & 37.76 & 0.0091 & 0.0272 \\
            & REDfull\_ReCoSa & 38.07 & 0.0073 &  0.0202 \\
    \bottomrule
    \end{tabular}%
\end{adjustbox}
\end{table}

\begin{table}
    \caption{Performance of queries of different order for five models on the JDC dataset and the bAbI dialog dataset.}
	\label{Table:exp_other}
	\begin{adjustbox}{max width=0.95\linewidth}
    \begin{tabular}{clccc}
    
    \multicolumn{5}{c}{JDC Dataset}\\
    \toprule
     & Models & PPL   & Dist-1 & Dist-2 \\
    \midrule
    \multirow{2}{*}{Seq2Seqs} 
            & Seq2Seq & 12.72 & 0.0035 & 0.0101 \\
            & REDfull\_Seq2Seq & \textbf{11.82} & \textbf{0.0037} & \textbf{0.0110} \\
    \midrule
    \multirow{2}{*}{SeqAtts} 
            & SeqAtt & 11.70 & \textbf{0.0086} & \textbf{0.0311} \\
            & REDfull\_SeqAtt & \textbf{11.68} & 0.0051 & 0.0178 \\
    \midrule
    \multirow{2}{*}{Transformers} 
            & Transformer & 10.40 & 0.0014 & 0.0027\\
            & REDfull\_Transformer & \textbf{10.28} & \textbf{0.0051} & \textbf{0.0140} \\
    \midrule
    \multirow{2}{*}{HREDs} 
            & HRED & 10.90 & 0.0087 & 0.0376 \\
            & REDfull\_HRED & \textbf{10.80} & \textbf{0.0091} & \textbf{0.0396} \\
    \midrule
    \multirow{2}{*}{ReCoSas} 
            & ReCoSa & 10.44 & 0.0015 & 0.0029\\
            & REDfull\_ReCoSa & \textbf{9.43} & \textbf{0.0111} & \textbf{0.0465} \\
    \bottomrule
    \\
    \multicolumn{5}{c}{bAbI Dataset}\\
    \toprule
     & Models & PPL   & Dist-1 & Dist-2 \\
    \midrule

    \multirow{2}{*}{Seq2Seqs} 
            & Seq2Seq & 1.21 &0.0007  & 0.0015 \\
            & REDfull\_Seq2Seq & \textbf{1.172}  & 0.0007 & \textbf{0.0016} \\
    \midrule
    \multirow{2}{*}{SeqAtts} 
            & SeqAtt & 1.032 & 0.0007 & 0.0020 \\
            & REDfull\_SeqAtt & \textbf{1.025} & 0.0007 & 0.0020 \\
    \midrule
    \multirow{2}{*}{Transformers} 
            & Transformer & 1.026 & 0.0007 & 0.0020 \\
            & REDfull\_Transformer & \textbf{1.023} & 0.0007 & 0.0020 \\
    \midrule
    \multirow{2}{*}{HREDs} 
            & HRED & 1.027 & 0.0007 & \textbf{0.0020} \\
            & REDfull\_HRED & \textbf{1.026}      &   0.0007    & 0.0019 \\
    \midrule
    \multirow{2}{*}{ReCoSas} 
            & ReCoSa & 1.024  & 0.0007 & 0.0019\\
            & REDfull\_ReCoSa & \textbf{1.019} & 0.0007 &  \textbf{0.0020} \\
    \bottomrule
    \end{tabular}%
\end{adjustbox}
\end{table}

\subsubsection{Evaluations}
We use both quantitative metrics and human judgments for evaluation in our experiments. Considering that BLEU~\cite{papineni2002bleu} and other word-overlap metrics correlate poorly with human judgements of response quality~\cite{liu2016not}, we use per-word perplexity (PPL)~\cite{bahl1983maximum} and Distinct~\cite{li2015diversity} as automatic metrics for quantitative comparisons. The traditional metric PPL describes how well our trained probabilistic model (denoted as $p$) predicts the expected ground-truth responses.
Given an utterance $u = \{w_1,w_2,\dots,w_n\}$, the PPL is defined as the exponentiation of the word entropy, 
\[
\textrm{PPL}(u)=\exp\bigl(H[p(u)]\bigr)=\exp\bigl(-\sum_{i=1}^Np(w_i)\log p(w_i)\bigr).
\]
The smaller the PPL, the closer the generated probabilistic distribution is to the ground-truth distribution. The metric PPL has been widely used in NLP and multi-turn dialogue generation \cite{chen2018hierarchical,tian2017make,xing2018hierarchical} to evaluate the quality of generated responses. 
We use Distinct to evaluate the degree of diversity of the generated responses by calculating the number of distinct unigrams and bigrams in the generated responses. The two numbers are scaled by the total
number of generated tokens and then denoted as dist-1 and dist-2. The larger the distinct values indicates the more informative and diverse the generated responses.

What's more, we also conduct a human evaluation to evaluate the ability of the models to generate good responses. Given 200 randomly sampled dialogue history turns, we choose the standard Transformer and ReCoSa, and their corresponding ranking enhanced RED1 and REDfull models (which model local and global dynamics respectively) to generate responses for each of the 200 sampled cases. We then invite 3 CS annotators to judge the quality of responses generated by the 6 representative models. Each of them is required to give the comparison between our proposed ranking enhanced models and the baseline standard models, e.g., win, loss, and tie, based on the informativeness, fluency and relevance with history. For example, the "win" label means that the generated response of model A is more proper than model B. Notably, the order of the responses is shuffled randomly.

\subsubsection{Baselines and Our Models}
We compare 5 baseline models, Seq2Seq, Seq2Seq with attention (SeqAtt), Transformer, HRED and ReCoSa, with those models applied our RED framework. 

For sequential models:
\begin{itemize}
    \item \textbf{Seq2Seq}: The RNN based sequence to sequence model uses LSTM for both encoder and decoder.
    \item \textbf{SeqAtt}: The se2seq with attention. It introduces the attention mechanism to the RNN-based Seq2Seq model.
    \item \textbf{Transformer}~\cite{vaswani2017attention}: The standard Transformer model which takes all of the history turns as a whole sequence of words.
\end{itemize}
The above three models are provided by the ParlAI~\cite{miller2017parlai} framework which is identical to the experiments in Sankar's empirical study~\cite{sankar2019neural}. The hyperparameters are exactly the same as which provided by their work for fair comparison on the bAbI and PersonaChat. 

For hierarchical models:
\begin{itemize}
    \item \textbf{HRED}~\cite{serban2016building}: The basic hierarchical model which uses a two-level hierarchical encoder to encode the history turns.
    \item \textbf{ReCoSa}~\cite{zhang2019recosa}: The state of the art hierarchical model which uses Transformer as the inter-utterance level encoder.
\end{itemize}
We implement HRED and ReCoSa in the ParlAI framework by ourselves, to reuse the standard components and the common process provided by the framework. 

\textbf{RED$\mathbf{x}$\_$\mathbf{y}$}: The prefix `RED' denotes our Ranking Enhanced Dialogue generative models, and the $\mathbf{x}$ in 1,2,or 3 denotes all of the ranking enhanced methods whose query are 1 order, 2 order or 3 order respectively. We set $\mathbf{x}$ to symbol `full' to denote ranking method who use all the history utterances as a query. The $\mathbf{y}$ denotes the baseline models from Seq2Seq, SeqAtt, Transformer, HRED or ReCoSa.
For example, RED2\_Transformer denotes the Transformer model with 2 order local ranking enhanced module.

The query is constructed using a single LSTM layer with 64 hidden size, and followed a fully connected layer with the hidden size of 128 in all RED based models. The hyperparameter $\alpha$ set to be 0.01 in loss function $\mathcal{L}$.
For the bAbI dialog and the PersonaChat datasets, in Seq2Seq, SeqAtt and their RED versions, the encoder and decoder components are 2-layer LSTM with 128 hidden size. 
For Transformer, we use 300 dimensional embeddings and hidden size, 2 layers and 2 attention heads. 
For HRED and related ranking enhanced models, we use a 2-layer LSTM with hidden size 128 as its utterance level encoder and use another 2-layer LSTM with hidden size 128 for both inter-utterance level encoder and decoder. 
For ReCoSa and related ranking enhanced models, we use a 2-layer LSTM with hidden size 300 as its utterance level encoder, the hidden size/layers/attention heads for both inter-utterance level Transformer encoder and Transformer decoder are 300/2/2 respectively.
For the JDC dataset, except for the LSTM encoder for constructing queries representations and the following fully connected layer in RED framework, all of the models on JDC use 512-dimensional hidden units and 512-dimensional word embedding both for LSTMs and Transformers. Additionally, the LSTM components reduce to a single-layer LSTM and the layers and attention heads for Transformers are 6 and 8 respectively.

Adam~\cite{kingma2014adam} is utilized for optimization, and the learning rate is set to be 0.005 for Seq2Seq/SeqAtt/HRED and their related ranking enhanced models, 0.001 for Transformer/ReCoSa and their related ranking enhanced models. We use early stopping with a patience of 10 on the validation set to obtain the best models.

The source code and all of the experiments can be found in
\url{https://github.com/ying-A/RED}

\subsection{Experimental Results}

\textbf{Metric-based Evaluation}
We do comparative experiments on the three datasets to observe the differences of quantitative metrics between the baseline models and our ranking enhanced models. Specifically, we calculate PPL and distinct on the test datasets to compare the generative ability of different models fairly. Results are shown in Tables ~\ref{Table:exp_personachat} and ~\ref{Table:exp_other}. On the three datasets, all RED models outperform baselines in PPL. For the bAbI dataset, the gaps in PPL are small, mainly because the task is relatively simpler compared with the other two datasets, even the basic Seq2Seq model can achieve a nearly perfect PPL close to 1. Thus the RED models also achieve almost the same distincts with baselines. For the JDC dataset and the PersonaChat dataset, our improvements on PPL are significant and the vast majority of RED models get higher distincts than baselines. Take the results on the PersonaChat dataset for example. For PPL, all of the RED models in the five groups achieve lower PPL than their related baseline models. It reveals that by additionally learning a task of ranking history utterances, the generated probabilities of the generative module can be closer to the ground-truth distribution in the probability distribution space. For distinct, we find that the highest distinct values are all got by our RED models except for models in Transformer groups. It shows that the RED models are better for generating diverse responses by explicitly modeling dynamics in history turns.

Moreover, RED3\_Seq2Seq, RED3\_SeqAtt, RED2\_Transformer, RED2\_HRED and RED2\_ReCoSa achieve the lowest PPL in each of their related group, REDfull\_Transformer and REDfull\_HRED achieve the second-lowest PPL in each of their related group. It shows that both local ranking methods and global ranking methods are helpful for improving generative performance. We also find what matters most for improving generative performance is the ability to model sequential order information for original baseline models. As the ability to model the sequential dynamic flow for original baseline models becomes weaker and weaker, the performance improvements (PPL decreases) brought by RED models become bigger and bigger. The largest PPL decreases for baselines Seq2Seq(0.34), SeqAtt(0.42), ReCoSa(0.7), HRED(0.95), Transformer(1.03) respectively. Take a deep look into the phenomenon, RNN structures are inherently capable to model sequential order. Seq2Seq and SeqAtt use fully RNN structures to model word order information. Hierarchical models HRED and ReCoSa also use an utterance level RNN to model word order information, however, the word order information is separately modeled in each individual utterance.  While the Transformer only uses a weak positional embedding to keep order information, thus the improvements for Transformer are larger than other baselines.

\textbf{Human Evaluation and Case Study}
For qualitative evaluation, we use human evaluation to compare responses generated by different models. The results are shown in Table~\ref{human evaluation}. The percentage of win, loss and tie, as compared with the baselines in terms of informativeness/fluency/relevance, are given to evaluate the quality of generated responses by RED models. The results show that all of the RED models achieve better performances (\#win $-$ \#loss) in the three aspects, which reveals that our RED models generate responses with higher quality than baselines. Kappa~\cite{fleiss1971measuring} values are presented to demonstrate the consistency of different annotators. 

Table~\ref{case study} presents two samples generated by RED models and baselines. For the first example, person A first answer that he likes the music by `John Denver', and ask what B likes by an elliptical sentence `how about you'. We can see that baselines Transformer and ReCoSa only care about the utterance `how about you' without correctly understanding the logical relationship between it and its previous history utterances, and thus generate wrong answers. However, our RED models generate coherent answers concentrating on the current dialogue topic music. For the second example, the utterance of the last history turn is about pets and dogs. Transformer gives its reply by the word `brother' that appeared twice in past history utterances, and ReCoSa generates two sentences with inconsistent logic. Our RED models generate informative responses about the pets. The two examples show that baselines ignore the dynamic logical information in history turns, while RED models give appropriate answers by learning the dynamic history structure.

\subsection{Empirical Analysis}
We conducted additional experiments to analyze the reasons that the RED framework outperforms the baselines. The analysis was conducted using the results on the PersonaChat dataset as examples. The similar phenomenon is also observed on the other two datasets.

\subsubsection{\textbf{Where does the improvement comes from}}

To identify where the improvement for RED models comes from, we split the dialogue history into three groups of similar size according to their lengths, including length less than 11, between 11 and 15, and longer than 15 utterances.
The 968 multi-turn conversation instances in the PersonaChat dataset can be constructed into 6544 history-response pairs in total, in this way, three groups contain 1937/2431/2176 instances respectively.
The Transformer and its RED enhanced versions are evaluated on these new divided data partitions, and the results are shown in Table~\ref{Table:ana_where}.
For each model on each data partitions, we calculate their PPL and $\Delta$PPL metrics, where $\Delta$PPL is the difference between each RED model and original Transformer model under a specific history length. For example, $\Delta$PPL in the 5-th row equals to PPL in the 5-th row minus PPL in the 1-st row. $\Delta$PPL indicates the improvement of each RED model compares to the baseline under different history length data partitions.

As shown in Table~\ref{Table:ana_where}, in each block, PPL increases along with the increasing of history length for all models, reflecting that generating the response with longer history utterance is more difficult than the shorter one. Longer dialogue history brings not only more informative signals, but also more complex dynamics, which makes it harder for response generation. 
As for the $\Delta$PPL in each RED model, e.g., RED1\_Transformer, the values decrease along with the increasing of history length, e.g., from -0.46, -0.74 to -1.95. The phenomenon reveals that the RED framework, regardless of query orders, is more helpful for the instance with longer dialogue history by modeling the order of the utterances explicitly. It makes RED able to capture the dynamic flow easier than the original Transformer, especially for longer dialogue history which is a tougher part in the multi-turn dialogue generation task.

\begin{table}
    \caption{PPL of queries of different order for Transformer on PersonaChat split dataset with different history length.}
	\label{Table:ana_where}
    \begin{tabular}{cccc}
    \toprule
    Model & \multicolumn{1}{c}{History Length} & PPL & $\Delta$PPL \\
    \midrule
    \multirow{4}[3]{*}{Transformer} 
    		& <11 & 35.82 & \multirow{3}{*}{-} \\
          	& 11-15 & 40.37 &  \\
          	& >15   & 45.78 &  \\
    \cmidrule{2-4}      & all   & 40.48 &  \\
    \midrule
    \multirow{4}[2]{*}{RED1\_Transformer} 
    		& <11   & 35.36 & -0.46 \\
          	& 11-15 & 39.63 & -0.74 \\
          	& >15   & 43.83 & -1.95 \\
    \cmidrule{2-4}      & all   & 39.49 & -0.99 \\
    \midrule
    \multirow{4}[2]{*}{RED2\_Transformer} 
    		& <11   & 35.72 & -0.10 \\
          	& 11-15 & 39.47 & -0.90 \\
          	& >15   & 43.44 & -2.34 \\
    \cmidrule{2-4}      & all   & 39.45 & -1.03 \\
    \midrule
    \multirow{4}[2]{*}{RED3\_Transformer} 
    		& <11   & 35.67 & -0.15 \\
          	& 11-15 & 39.41 & -0.96 \\
          	& >15   & 43.88 & -1.90 \\
    \cmidrule{2-4}      & all   & 39.54 & -0.94 \\
    \midrule
    \multirow{4}[3]{*}{REDfull\_Transformer} 
    		& <11   & 35.33 & -0.49 \\
          	& 11-15 & 39.52 & -0.85 \\
          	& >15   & 43.88 & -1.90 \\
    \cmidrule{2-4}      & all   & 39.46 & -1.02 \\
    \bottomrule
    \end{tabular}%
\end{table}

\begin{figure} 
	\begin{minipage}[t]{0.48\linewidth}
		\centerline{\includegraphics[width=1\linewidth]{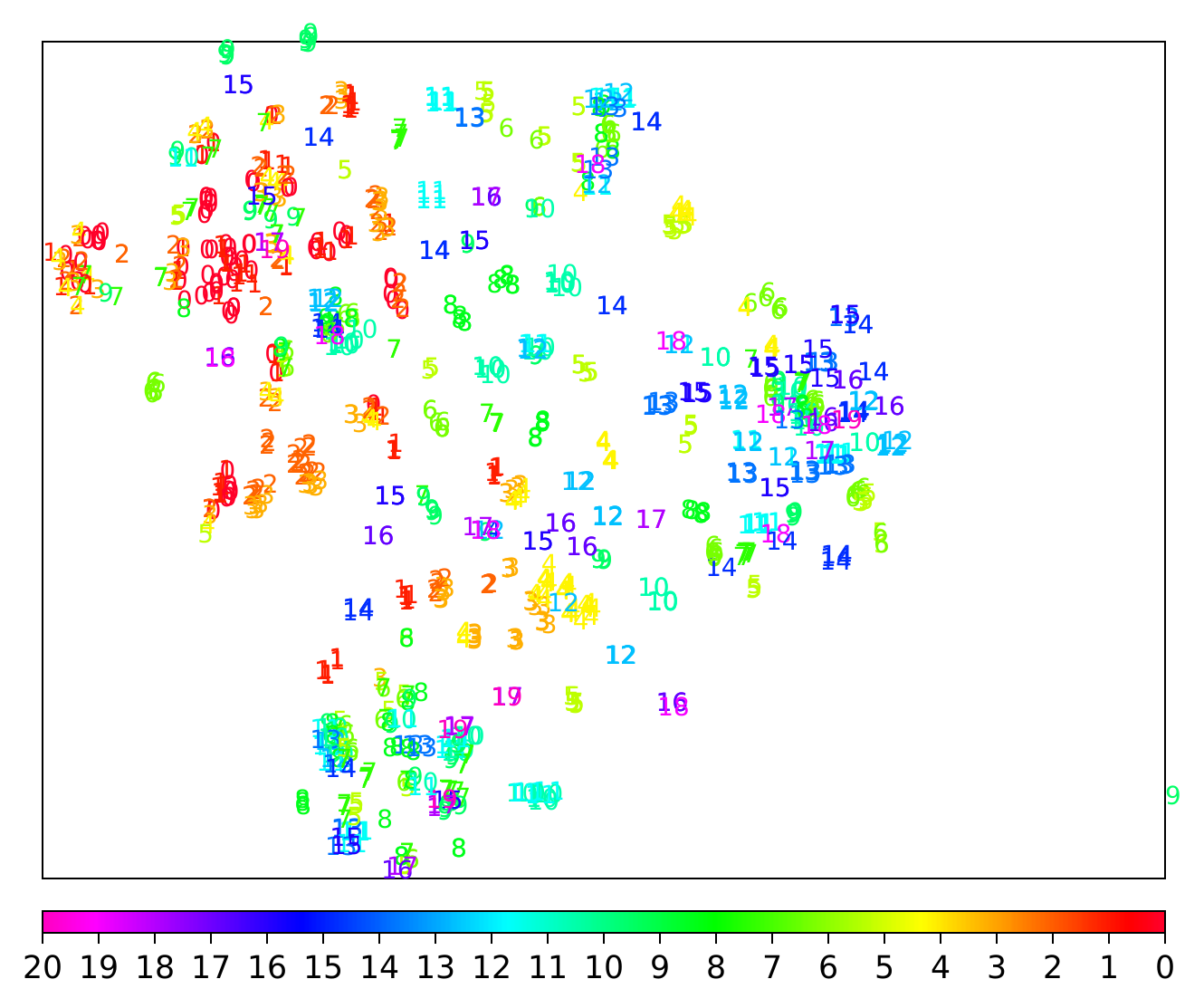}}
		\caption*{(A) Transformer}
	\end{minipage}%
	\begin{minipage}[t]{0.48\linewidth}
		\centerline{\includegraphics[width=1\linewidth]{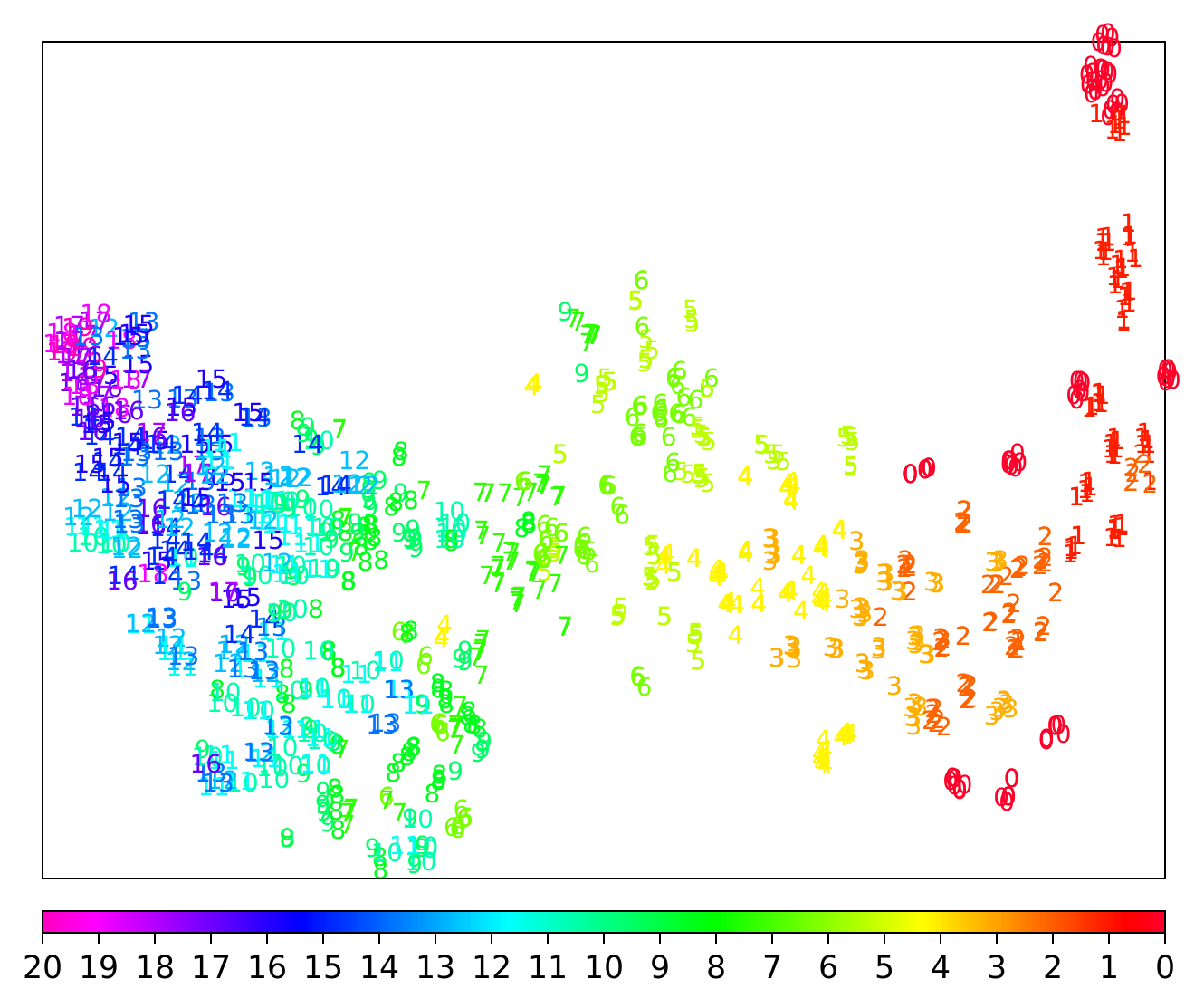}}
		\caption*{(B) REDfull\_Transformer}
	\end{minipage}
    \caption{ t-SNE embeddings of representations learned by the models for utterances from the PersonaChat test set. Embeddings are color coded by the position of the utterance in the dialogue history it appears.}\label{Figure:ana_how}
\end{figure}

\begin{figure}
    \centering
    \includegraphics[width=0.92\linewidth]{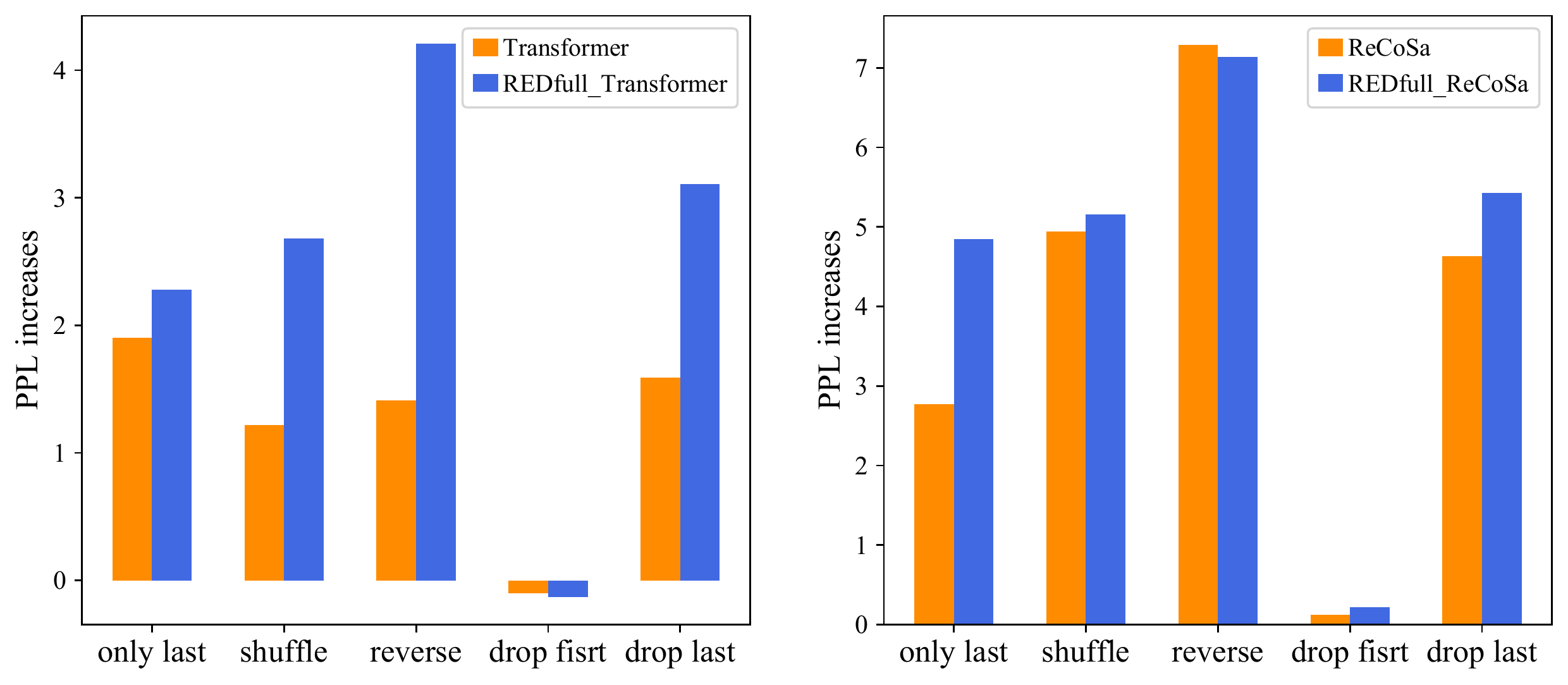}
    \caption{PPL increases for 4 models (Left: Transformer and REDfull\_Transformer, Right: ReCoSa and REDfull\_ReCoSa) when applying 5 kinds of perturbations. }
    \label{fig:perturb}
\end{figure}
\subsubsection{\textbf{How does the improvement comes from}}

The ranking loss for RED models is indeed a type of regularization, and it demands the models learn to rank by learning better utterance representations. Meanwhile, the enhanced representations can help generation module effectively use the dynamic information encoded in them and generate better responses. To study whether RED representations encoded dynamics well, we analyze the learned utterance representations for RED models and baseline models.

Without loss of generality, we use the trained Transformer and REDfull\_Transformer as our two comparison models, and analyze the representations of utterances on the PersonaChat test dataset.
For each model, we random sample 1000 representations of utterances with different positional-indexes (1 to max history length) in the history. After that, an off-the-shelf technique - t-SNE~\cite{maaten2008visualizing} is used to reduce the representation dimensions for further visualization. Specifically, we reduce the representations to 2-dimensional vectors and draw them with positional-indexes in an x-y coordinate.

It is evident that in REDfull\_Transformer, see Figure~\ref{Figure:ana_how}(B), the vectors with the same positional-index gather in a tight cluster. The clusters form a clear sense of hierarchy according to their positional orders, vividly showing a scene of dynamic flow. However, such structure information is not illustrated in Transformer as in Figure~\ref{Figure:ana_how}(A), where variant labeled vectors are scattered everywhere. 
It indicates that RED learned the high-level dynamic structures by encoding this information in the utterance representations, while traditional models do not have sufficient ability to do that.

What's more, we also do utterance perturbation experiments on the PersonaChat test set as implemented in Sankar's empirical study\cite{sankar2019neural}. We first get the PPL for different models on the test set. Then we perturb the dialogue history and record the PPL on the perturbed test set. The PPL increases after perturbations for all of the models are calculated and shown in Figure~\ref{fig:perturb}. 
Results show that all of the REDfull models get larger PPL increases in most kinds of perturbations than origin models and thus more sensitive to history utterance perturbations. It proves that the dynamic (order) information is more effectively used by RED models according to the premise in \cite{sankar2019neural} that \textit{the more sensitive the model to perturbations, the stronger ability for it of modeling dynamics}.

\section{Conclusion}
In this paper, we proposed a Ranking Enhanced Dialogue generation framework, namely RED, to explicitly capture the dynamics in the dialogue history. RED consists of the regular representation encoder module and response generation module, that have been used in the original dialogue generation models, and an additionally novel ranking enhanced module. In the ranking module, a learning-to-rank task is designed for ranking dialogue history, where the former utterance is treated as a query, and the consecutive utterances after the query are treated as the documents. RED is a general framework that can easily cooperate with most of the recent multi-turn dialogue generation models. With the help of RED, we observe the improvement in terms of both quantitative metrics and human evaluations. Further empirical analysis demonstrates that the RED is especially good at tackling longer dialogue history, by encoding history dynamics into the utterance representations.

\begin{acks}
This work was supported by the National Key R\&D Program of China under Grants No. 2019AAA0105200, 2016QY02D0405, the Beijing Academy of Artificial Intelligence (BAAI) (No. BAAI2019ZD0306, BAAI2020ZJ0303), the National Natural Science Foundation of China (NSFC) (No. 61722211, 61773362, 61872338, 61906180), the Lenovo-CAS Joint Lab Youth Scientist Project, the Foundation and Frontier Research Key Program of Chongqing Science and Technology Commission (No. cstc2017jcyjBX0059), the Tencent AI Lab Rhino-Bird Focused Research Program (No. JR202033), and the K.C.Wong Education Foundation.
\end{acks}

\balance
\bibliographystyle{ACM-Reference-Format}
\bibliography{CIKM2020_RED}


\begin{thebibliography}{36}


\ifx \showCODEN    \undefined \def \showCODEN     #1{\unskip}     \fi
\ifx \showDOI      \undefined \def \showDOI       #1{#1}\fi
\ifx \showISBNx    \undefined \def \showISBNx     #1{\unskip}     \fi
\ifx \showISBNxiii \undefined \def \showISBNxiii  #1{\unskip}     \fi
\ifx \showISSN     \undefined \def \showISSN      #1{\unskip}     \fi
\ifx \showLCCN     \undefined \def \showLCCN      #1{\unskip}     \fi
\ifx \shownote     \undefined \def \shownote      #1{#1}          \fi
\ifx \showarticletitle \undefined \def \showarticletitle #1{#1}   \fi
\ifx \showURL      \undefined \def \showURL       {\relax}        \fi
\providecommand\bibfield[2]{#2}
\providecommand\bibinfo[2]{#2}
\providecommand\natexlab[1]{#1}
\providecommand\showeprint[2][]{arXiv:#2}

\bibitem[\protect\citeauthoryear{Bahdanau, Cho, and Bengio}{Bahdanau
  et~al\mbox{.}}{2014}]%
        {bahdanau2014neural}
\bibfield{author}{\bibinfo{person}{Dzmitry Bahdanau},
  \bibinfo{person}{Kyunghyun Cho}, {and} \bibinfo{person}{Yoshua Bengio}.}
  \bibinfo{year}{2014}\natexlab{}.
\newblock \showarticletitle{Neural machine translation by jointly learning to
  align and translate}.
\newblock \bibinfo{journal}{\emph{arXiv preprint arXiv:1409.0473}}
  (\bibinfo{year}{2014}).
\newblock


\bibitem[\protect\citeauthoryear{Bahl, Jelinek, and Mercer}{Bahl
  et~al\mbox{.}}{1983}]%
        {bahl1983maximum}
\bibfield{author}{\bibinfo{person}{Lalit~R Bahl}, \bibinfo{person}{Frederick
  Jelinek}, {and} \bibinfo{person}{Robert~L Mercer}.}
  \bibinfo{year}{1983}\natexlab{}.
\newblock \showarticletitle{A maximum likelihood approach to continuous speech
  recognition}.
\newblock \bibinfo{journal}{\emph{IEEE transactions on pattern analysis and
  machine intelligence}} \bibinfo{number}{2} (\bibinfo{year}{1983}),
  \bibinfo{pages}{179--190}.
\newblock


\bibitem[\protect\citeauthoryear{Bengio, Simard, Frasconi,
  et~al\mbox{.}}{Bengio et~al\mbox{.}}{1994}]%
        {bengio1994learning}
\bibfield{author}{\bibinfo{person}{Yoshua Bengio}, \bibinfo{person}{Patrice
  Simard}, \bibinfo{person}{Paolo Frasconi}, {et~al\mbox{.}}}
  \bibinfo{year}{1994}\natexlab{}.
\newblock \showarticletitle{Learning long-term dependencies with gradient
  descent is difficult}.
\newblock \bibinfo{journal}{\emph{IEEE transactions on neural networks}}
  \bibinfo{volume}{5}, \bibinfo{number}{2} (\bibinfo{year}{1994}),
  \bibinfo{pages}{157--166}.
\newblock


\bibitem[\protect\citeauthoryear{Bordes, Boureau, and Weston}{Bordes
  et~al\mbox{.}}{2016}]%
        {bordes2016learning}
\bibfield{author}{\bibinfo{person}{Antoine Bordes}, \bibinfo{person}{Y-Lan
  Boureau}, {and} \bibinfo{person}{Jason Weston}.}
  \bibinfo{year}{2016}\natexlab{}.
\newblock \showarticletitle{Learning end-to-end goal-oriented dialog}.
\newblock \bibinfo{journal}{\emph{arXiv preprint arXiv:1605.07683}}
  (\bibinfo{year}{2016}).
\newblock


\bibitem[\protect\citeauthoryear{Chen, Ren, Tang, Zhao, and Yin}{Chen
  et~al\mbox{.}}{2018}]%
        {chen2018hierarchical}
\bibfield{author}{\bibinfo{person}{Hongshen Chen}, \bibinfo{person}{Zhaochun
  Ren}, \bibinfo{person}{Jiliang Tang}, \bibinfo{person}{Yihong~Eric Zhao},
  {and} \bibinfo{person}{Dawei Yin}.} \bibinfo{year}{2018}\natexlab{}.
\newblock \showarticletitle{Hierarchical variational memory network for
  dialogue generation}. In \bibinfo{booktitle}{\emph{Proceedings of the 2018
  WWW Conference}}. \bibinfo{pages}{1653--1662}.
\newblock


\bibitem[\protect\citeauthoryear{Eric and Manning}{Eric and Manning}{2017}]%
        {eric2017key}
\bibfield{author}{\bibinfo{person}{Mihail Eric} {and}
  \bibinfo{person}{Christopher~D Manning}.} \bibinfo{year}{2017}\natexlab{}.
\newblock \showarticletitle{Key-value retrieval networks for task-oriented
  dialogue}.
\newblock \bibinfo{journal}{\emph{arXiv preprint arXiv:1705.05414}}
  (\bibinfo{year}{2017}).
\newblock


\bibitem[\protect\citeauthoryear{Fleiss}{Fleiss}{1971}]%
        {fleiss1971measuring}
\bibfield{author}{\bibinfo{person}{Joseph~L Fleiss}.}
  \bibinfo{year}{1971}\natexlab{}.
\newblock \showarticletitle{Measuring nominal scale agreement among many
  raters.}
\newblock \bibinfo{journal}{\emph{Psychological bulletin}}
  \bibinfo{volume}{76}, \bibinfo{number}{5} (\bibinfo{year}{1971}),
  \bibinfo{pages}{378}.
\newblock


\bibitem[\protect\citeauthoryear{Kim, Banchs, and Li}{Kim
  et~al\mbox{.}}{2016}]%
        {kim2016exploring}
\bibfield{author}{\bibinfo{person}{Seokhwan Kim}, \bibinfo{person}{Rafael~E
  Banchs}, {and} \bibinfo{person}{Haizhou Li}.}
  \bibinfo{year}{2016}\natexlab{}.
\newblock \showarticletitle{Exploring convolutional and recurrent neural
  networks in sequential labelling for dialogue topic tracking}. In
  \bibinfo{booktitle}{\emph{Proceedings of the 54th Annual Meeting of the ACL
  (Volume 1: Long Papers)}}. \bibinfo{pages}{963--973}.
\newblock


\bibitem[\protect\citeauthoryear{Kingma and Ba}{Kingma and Ba}{2014}]%
        {kingma2014adam}
\bibfield{author}{\bibinfo{person}{Diederik~P Kingma} {and}
  \bibinfo{person}{Jimmy Ba}.} \bibinfo{year}{2014}\natexlab{}.
\newblock \showarticletitle{Adam: A method for stochastic optimization}.
\newblock \bibinfo{journal}{\emph{arXiv preprint arXiv:1412.6980}}
  (\bibinfo{year}{2014}).
\newblock


\bibitem[\protect\citeauthoryear{Lee, Prasad, Joshi, Dinesh, and Webber}{Lee
  et~al\mbox{.}}{2006}]%
        {lee2006complexity}
\bibfield{author}{\bibinfo{person}{Alan Lee}, \bibinfo{person}{Rashmi Prasad},
  \bibinfo{person}{Aravind Joshi}, \bibinfo{person}{Nikhil Dinesh}, {and}
  \bibinfo{person}{Bonnie Webber}.} \bibinfo{year}{2006}\natexlab{}.
\newblock \showarticletitle{Complexity of dependencies in discourse: Are
  dependencies in discourse more complex than in syntax}. In
  \bibinfo{booktitle}{\emph{Proceedings of the 5th International Workshop on
  Treebanks and Linguistic Theories}}. \bibinfo{pages}{12--23}.
\newblock


\bibitem[\protect\citeauthoryear{Li, Galley, Brockett, Gao, and Dolan}{Li
  et~al\mbox{.}}{2015}]%
        {li2015diversity}
\bibfield{author}{\bibinfo{person}{Jiwei Li}, \bibinfo{person}{Michel Galley},
  \bibinfo{person}{Chris Brockett}, \bibinfo{person}{Jianfeng Gao}, {and}
  \bibinfo{person}{Bill Dolan}.} \bibinfo{year}{2015}\natexlab{}.
\newblock \showarticletitle{A diversity-promoting objective function for neural
  conversation models}.
\newblock \bibinfo{journal}{\emph{arXiv preprint arXiv:1510.03055}}
  (\bibinfo{year}{2015}).
\newblock


\bibitem[\protect\citeauthoryear{Li, Su, Shen, Li, Cao, and Niu}{Li
  et~al\mbox{.}}{2017}]%
        {li2017dailydialog}
\bibfield{author}{\bibinfo{person}{Yanran Li}, \bibinfo{person}{Hui Su},
  \bibinfo{person}{Xiaoyu Shen}, \bibinfo{person}{Wenjie Li},
  \bibinfo{person}{Ziqiang Cao}, {and} \bibinfo{person}{Shuzi Niu}.}
  \bibinfo{year}{2017}\natexlab{}.
\newblock \showarticletitle{DailyDialog: A Manually Labelled Multi-turn
  Dialogue Dataset}. In \bibinfo{booktitle}{\emph{Proceedings of the Eighth
  IJCNLP (Volume 1: Long Papers)}}. \bibinfo{pages}{986--995}.
\newblock


\bibitem[\protect\citeauthoryear{Liu, Lowe, Serban, Noseworthy, Charlin, and
  Pineau}{Liu et~al\mbox{.}}{2016}]%
        {liu2016not}
\bibfield{author}{\bibinfo{person}{Chia-Wei Liu}, \bibinfo{person}{Ryan Lowe},
  \bibinfo{person}{Iulian~Vlad Serban}, \bibinfo{person}{Mike Noseworthy},
  \bibinfo{person}{Laurent Charlin}, {and} \bibinfo{person}{Joelle Pineau}.}
  \bibinfo{year}{2016}\natexlab{}.
\newblock \showarticletitle{How NOT To Evaluate Your Dialogue System: An
  Empirical Study of Unsupervised Evaluation Metrics for Dialogue Response
  Generation}. In \bibinfo{booktitle}{\emph{Proceedings of the 2016 Conference
  on EMNLP}}. \bibinfo{pages}{2122--2132}.
\newblock


\bibitem[\protect\citeauthoryear{Liu et~al\mbox{.}}{Liu et~al\mbox{.}}{2009}]%
        {liu2009learning}
\bibfield{author}{\bibinfo{person}{Tie-Yan Liu} {et~al\mbox{.}}}
  \bibinfo{year}{2009}\natexlab{}.
\newblock \showarticletitle{Learning to rank for information retrieval}.
\newblock \bibinfo{journal}{\emph{Foundations and Trends{\textregistered} in
  Information Retrieval}} \bibinfo{volume}{3}, \bibinfo{number}{3}
  (\bibinfo{year}{2009}), \bibinfo{pages}{225--331}.
\newblock


\bibitem[\protect\citeauthoryear{Lowe, Pow, Serban, and Pineau}{Lowe
  et~al\mbox{.}}{2015}]%
        {lowe2015ubuntu}
\bibfield{author}{\bibinfo{person}{Ryan Lowe}, \bibinfo{person}{Nissan Pow},
  \bibinfo{person}{Iulian Serban}, {and} \bibinfo{person}{Joelle Pineau}.}
  \bibinfo{year}{2015}\natexlab{}.
\newblock \showarticletitle{The Ubuntu Dialogue Corpus: A Large Dataset for
  Research in Unstructured Multi-Turn Dialogue Systems}. In
  \bibinfo{booktitle}{\emph{Proceedings of the 16th Annual Meeting of the
  SIGDIAL}}. \bibinfo{pages}{285--294}.
\newblock


\bibitem[\protect\citeauthoryear{Maaten and Hinton}{Maaten and Hinton}{2008}]%
        {maaten2008visualizing}
\bibfield{author}{\bibinfo{person}{Laurens van~der Maaten} {and}
  \bibinfo{person}{Geoffrey Hinton}.} \bibinfo{year}{2008}\natexlab{}.
\newblock \showarticletitle{Visualizing data using t-SNE}.
\newblock \bibinfo{journal}{\emph{JMLR}} \bibinfo{volume}{9},
  \bibinfo{number}{Nov} (\bibinfo{year}{2008}), \bibinfo{pages}{2579--2605}.
\newblock


\bibitem[\protect\citeauthoryear{Mikolov, Karafi{\'a}t, Burget,
  {\v{C}}ernock{\`y}, and Khudanpur}{Mikolov et~al\mbox{.}}{2010}]%
        {mikolov2010recurrent}
\bibfield{author}{\bibinfo{person}{Tom{\'a}{\v{s}} Mikolov},
  \bibinfo{person}{Martin Karafi{\'a}t}, \bibinfo{person}{Luk{\'a}{\v{s}}
  Burget}, \bibinfo{person}{Jan {\v{C}}ernock{\`y}}, {and}
  \bibinfo{person}{Sanjeev Khudanpur}.} \bibinfo{year}{2010}\natexlab{}.
\newblock \showarticletitle{Recurrent neural network based language model}. In
  \bibinfo{booktitle}{\emph{Eleventh annual conference of the ISCA}}.
\newblock


\bibitem[\protect\citeauthoryear{{Miller}, {Feng}, {Fisch}, {Lu}, {Batra},
  {Bordes}, {Parikh}, and {Weston}}{{Miller} et~al\mbox{.}}{2017}]%
        {miller2017parlai}
\bibfield{author}{\bibinfo{person}{A.~H. {Miller}}, \bibinfo{person}{W.
  {Feng}}, \bibinfo{person}{A. {Fisch}}, \bibinfo{person}{J. {Lu}},
  \bibinfo{person}{D. {Batra}}, \bibinfo{person}{A. {Bordes}},
  \bibinfo{person}{D. {Parikh}}, {and} \bibinfo{person}{J. {Weston}}.}
  \bibinfo{year}{2017}\natexlab{}.
\newblock \showarticletitle{ParlAI: A Dialog Research Software Platform}.
\newblock \bibinfo{journal}{\emph{arXiv preprint arXiv:{1705.06476}}}
  (\bibinfo{year}{2017}).
\newblock


\bibitem[\protect\citeauthoryear{Papineni, Roukos, Ward, and Zhu}{Papineni
  et~al\mbox{.}}{2002}]%
        {papineni2002bleu}
\bibfield{author}{\bibinfo{person}{Kishore Papineni}, \bibinfo{person}{Salim
  Roukos}, \bibinfo{person}{Todd Ward}, {and} \bibinfo{person}{Wei-Jing Zhu}.}
  \bibinfo{year}{2002}\natexlab{}.
\newblock \showarticletitle{BLEU: a method for automatic evaluation of machine
  translation}. In \bibinfo{booktitle}{\emph{Proceedings of the 40th annual
  meeting on ACL}}. \bibinfo{pages}{311--318}.
\newblock


\bibitem[\protect\citeauthoryear{Peng, Fang, Xie, and Zhou}{Peng
  et~al\mbox{.}}{2019}]%
        {peng2019topic}
\bibfield{author}{\bibinfo{person}{Yehong Peng}, \bibinfo{person}{Yizhen Fang},
  \bibinfo{person}{Zhiwen Xie}, {and} \bibinfo{person}{Guangyou Zhou}.}
  \bibinfo{year}{2019}\natexlab{}.
\newblock \showarticletitle{Topic-enhanced emotional conversation generation
  with attention mechanism}.
\newblock \bibinfo{journal}{\emph{Knowledge-Based Systems}}
  \bibinfo{volume}{163} (\bibinfo{year}{2019}), \bibinfo{pages}{429--437}.
\newblock


\bibitem[\protect\citeauthoryear{Sankar, Subramanian, Pal, Chandar, and
  Bengio}{Sankar et~al\mbox{.}}{2019}]%
        {sankar2019neural}
\bibfield{author}{\bibinfo{person}{Chinnadhurai Sankar},
  \bibinfo{person}{Sandeep Subramanian}, \bibinfo{person}{Christopher Pal},
  \bibinfo{person}{Sarath Chandar}, {and} \bibinfo{person}{Yoshua Bengio}.}
  \bibinfo{year}{2019}\natexlab{}.
\newblock \showarticletitle{Do Neural Dialog Systems Use the Conversation
  History Effectively? An Empirical Study}.
\newblock \bibinfo{journal}{\emph{arXiv preprint arXiv:1906.01603}}
  (\bibinfo{year}{2019}).
\newblock


\bibitem[\protect\citeauthoryear{Serban, Sordoni, Bengio, Courville, and
  Pineau}{Serban et~al\mbox{.}}{2016}]%
        {serban2016building}
\bibfield{author}{\bibinfo{person}{Iulian~V Serban},
  \bibinfo{person}{Alessandro Sordoni}, \bibinfo{person}{Yoshua Bengio},
  \bibinfo{person}{Aaron Courville}, {and} \bibinfo{person}{Joelle Pineau}.}
  \bibinfo{year}{2016}\natexlab{}.
\newblock \showarticletitle{Building end-to-end dialogue systems using
  generative hierarchical neural network models}. In
  \bibinfo{booktitle}{\emph{Thirtieth AAAI Conference on Artificial
  Intelligence}}.
\newblock


\bibitem[\protect\citeauthoryear{Serban, Sordoni, Lowe, Charlin, Pineau,
  Courville, and Bengio}{Serban et~al\mbox{.}}{2017}]%
        {serban2017hierarchical}
\bibfield{author}{\bibinfo{person}{Iulian~Vlad Serban},
  \bibinfo{person}{Alessandro Sordoni}, \bibinfo{person}{Ryan Lowe},
  \bibinfo{person}{Laurent Charlin}, \bibinfo{person}{Joelle Pineau},
  \bibinfo{person}{Aaron Courville}, {and} \bibinfo{person}{Yoshua Bengio}.}
  \bibinfo{year}{2017}\natexlab{}.
\newblock \showarticletitle{A hierarchical latent variable encoder-decoder
  model for generating dialogues}. In \bibinfo{booktitle}{\emph{Thirty-First
  AAAI Conference on Artificial Intelligence}}.
\newblock


\bibitem[\protect\citeauthoryear{Sutskever, Vinyals, and Le}{Sutskever
  et~al\mbox{.}}{2014}]%
        {sutskever2014sequence}
\bibfield{author}{\bibinfo{person}{Ilya Sutskever}, \bibinfo{person}{Oriol
  Vinyals}, {and} \bibinfo{person}{Quoc~V Le}.}
  \bibinfo{year}{2014}\natexlab{}.
\newblock \showarticletitle{Sequence to sequence learning with neural
  networks}. In \bibinfo{booktitle}{\emph{Advance in neurIPS}}.
  \bibinfo{pages}{3104--3112}.
\newblock


\bibitem[\protect\citeauthoryear{Tian, Yan, Mou, Song, Feng, and Zhao}{Tian
  et~al\mbox{.}}{2017}]%
        {tian2017make}
\bibfield{author}{\bibinfo{person}{Zhiliang Tian}, \bibinfo{person}{Rui Yan},
  \bibinfo{person}{Lili Mou}, \bibinfo{person}{Yiping Song},
  \bibinfo{person}{Yansong Feng}, {and} \bibinfo{person}{Dongyan Zhao}.}
  \bibinfo{year}{2017}\natexlab{}.
\newblock \showarticletitle{How to make context more useful? an empirical study
  on context-aware neural conversational models}. In
  \bibinfo{booktitle}{\emph{Proceedings of the 55th Annual Meeting of the ACL
  (Volume 2: Short Papers)}}. \bibinfo{pages}{231--236}.
\newblock


\bibitem[\protect\citeauthoryear{Traum and Heeman}{Traum and Heeman}{1996}]%
        {traum1996utterance}
\bibfield{author}{\bibinfo{person}{David~R Traum} {and}
  \bibinfo{person}{Peter~A Heeman}.} \bibinfo{year}{1996}\natexlab{}.
\newblock \showarticletitle{Utterance units in spoken dialogue}. In
  \bibinfo{booktitle}{\emph{Workshop on Dialogue Processing in Spoken Language
  Systems}}. Springer, \bibinfo{pages}{125--140}.
\newblock


\bibitem[\protect\citeauthoryear{Vaswani, Shazeer, Parmar, Uszkoreit, Jones,
  Gomez, Kaiser, and Polosukhin}{Vaswani et~al\mbox{.}}{2017}]%
        {vaswani2017attention}
\bibfield{author}{\bibinfo{person}{Ashish Vaswani}, \bibinfo{person}{Noam
  Shazeer}, \bibinfo{person}{Niki Parmar}, \bibinfo{person}{Jakob Uszkoreit},
  \bibinfo{person}{Llion Jones}, \bibinfo{person}{Aidan~N Gomez},
  \bibinfo{person}{{\L}ukasz Kaiser}, {and} \bibinfo{person}{Illia
  Polosukhin}.} \bibinfo{year}{2017}\natexlab{}.
\newblock \showarticletitle{Attention is all you need}. In
  \bibinfo{booktitle}{\emph{Advance in neurIPS}}. \bibinfo{pages}{5998--6008}.
\newblock


\bibitem[\protect\citeauthoryear{Vinyals and Le}{Vinyals and Le}{2015}]%
        {vinyals2015neural}
\bibfield{author}{\bibinfo{person}{Oriol Vinyals} {and} \bibinfo{person}{Quoc
  Le}.} \bibinfo{year}{2015}\natexlab{}.
\newblock \showarticletitle{A neural conversational model}.
\newblock \bibinfo{journal}{\emph{arXiv preprint arXiv:1506.05869}}
  (\bibinfo{year}{2015}).
\newblock


\bibitem[\protect\citeauthoryear{Wei, Liu, Mao, Guo, Zhu, Zhou, and Hu}{Wei
  et~al\mbox{.}}{2019}]%
        {wei2019emotion}
\bibfield{author}{\bibinfo{person}{Wei Wei}, \bibinfo{person}{Jiayi Liu},
  \bibinfo{person}{Xianling Mao}, \bibinfo{person}{Guibing Guo},
  \bibinfo{person}{Feida Zhu}, \bibinfo{person}{Pan Zhou}, {and}
  \bibinfo{person}{Yuchong Hu}.} \bibinfo{year}{2019}\natexlab{}.
\newblock \showarticletitle{Emotion-aware Chat Machine: Automatic Emotional
  Response Generation for Human-like Emotional Interaction}. In
  \bibinfo{booktitle}{\emph{Proceedings of the 28th ACM CIKM}}.
  \bibinfo{pages}{1401--1410}.
\newblock


\bibitem[\protect\citeauthoryear{Wen, Vandyke, Mrksic, Gasic, Rojas-Barahona,
  Su, Ultes, and Young}{Wen et~al\mbox{.}}{2016}]%
        {wen2016network}
\bibfield{author}{\bibinfo{person}{Tsung-Hsien Wen}, \bibinfo{person}{David
  Vandyke}, \bibinfo{person}{Nikola Mrksic}, \bibinfo{person}{Milica Gasic},
  \bibinfo{person}{Lina~M Rojas-Barahona}, \bibinfo{person}{Pei-Hao Su},
  \bibinfo{person}{Stefan Ultes}, {and} \bibinfo{person}{Steve Young}.}
  \bibinfo{year}{2016}\natexlab{}.
\newblock \showarticletitle{A network-based end-to-end trainable task-oriented
  dialogue system}.
\newblock \bibinfo{journal}{\emph{arXiv preprint arXiv:1604.04562}}
  (\bibinfo{year}{2016}).
\newblock


\bibitem[\protect\citeauthoryear{Xia, Liu, and Li}{Xia et~al\mbox{.}}{2009}]%
        {xia2009top}
\bibfield{author}{\bibinfo{person}{Fen Xia}, \bibinfo{person}{Tie-Yan Liu},
  {and} \bibinfo{person}{Hang Li}.} \bibinfo{year}{2009}\natexlab{}.
\newblock \showarticletitle{Top-k consistency of learning to rank methods}.
\newblock \bibinfo{journal}{\emph{Advance in NeurIPS}}  \bibinfo{volume}{22}
  (\bibinfo{year}{2009}), \bibinfo{pages}{2098--2106}.
\newblock


\bibitem[\protect\citeauthoryear{Xing, Wu, Wu, Liu, Huang, Zhou, and Ma}{Xing
  et~al\mbox{.}}{2017}]%
        {xing2017topic}
\bibfield{author}{\bibinfo{person}{Chen Xing}, \bibinfo{person}{Wei Wu},
  \bibinfo{person}{Yu Wu}, \bibinfo{person}{Jie Liu}, \bibinfo{person}{Yalou
  Huang}, \bibinfo{person}{Ming Zhou}, {and} \bibinfo{person}{Wei-Ying Ma}.}
  \bibinfo{year}{2017}\natexlab{}.
\newblock \showarticletitle{Topic aware neural response generation}. In
  \bibinfo{booktitle}{\emph{Thirty-First AAAI Conference on Artificial
  Intelligence}}.
\newblock


\bibitem[\protect\citeauthoryear{Xing, Wu, Wu, Huang, and Zhou}{Xing
  et~al\mbox{.}}{2018}]%
        {xing2018hierarchical}
\bibfield{author}{\bibinfo{person}{Chen Xing}, \bibinfo{person}{Yu Wu},
  \bibinfo{person}{Wei Wu}, \bibinfo{person}{Yalou Huang}, {and}
  \bibinfo{person}{Ming Zhou}.} \bibinfo{year}{2018}\natexlab{}.
\newblock \showarticletitle{Hierarchical recurrent attention network for
  response generation}. In \bibinfo{booktitle}{\emph{Thirty-Second AAAI
  Conference on Artificial Intelligence}}.
\newblock


\bibitem[\protect\citeauthoryear{Zhang, Lan, Pang, Guo, and Cheng}{Zhang
  et~al\mbox{.}}{2019}]%
        {zhang2019recosa}
\bibfield{author}{\bibinfo{person}{Hainan Zhang}, \bibinfo{person}{Yanyan Lan},
  \bibinfo{person}{Liang Pang}, \bibinfo{person}{Jiafeng Guo}, {and}
  \bibinfo{person}{Xueqi Cheng}.} \bibinfo{year}{2019}\natexlab{}.
\newblock \showarticletitle{ReCoSa: Detecting the Relevant Contexts with
  Self-Attention for Multi-turn Dialogue Generation}.
\newblock \bibinfo{journal}{\emph{arXiv preprint arXiv:1907.05339}}
  (\bibinfo{year}{2019}).
\newblock


\bibitem[\protect\citeauthoryear{Zhang, Dinan, Urbanek, Szlam, Kiela, and
  Weston}{Zhang et~al\mbox{.}}{2018}]%
        {zhang2018personalizing}
\bibfield{author}{\bibinfo{person}{Saizheng Zhang}, \bibinfo{person}{Emily
  Dinan}, \bibinfo{person}{Jack Urbanek}, \bibinfo{person}{Arthur Szlam},
  \bibinfo{person}{Douwe Kiela}, {and} \bibinfo{person}{Jason Weston}.}
  \bibinfo{year}{2018}\natexlab{}.
\newblock \showarticletitle{Personalizing Dialogue Agents: I have a dog, do you
  have pets too?}
\newblock \bibinfo{journal}{\emph{arXiv preprint arXiv:1801.07243}}
  (\bibinfo{year}{2018}).
\newblock


\bibitem[\protect\citeauthoryear{Zhou, Li, Dong, Liu, Chen, Zhao, Yu, and
  Wu}{Zhou et~al\mbox{.}}{2018}]%
        {zhou2018multi}
\bibfield{author}{\bibinfo{person}{Xiangyang Zhou}, \bibinfo{person}{Lu Li},
  \bibinfo{person}{Daxiang Dong}, \bibinfo{person}{Yi Liu},
  \bibinfo{person}{Ying Chen}, \bibinfo{person}{Wayne~Xin Zhao},
  \bibinfo{person}{Dianhai Yu}, {and} \bibinfo{person}{Hua Wu}.}
  \bibinfo{year}{2018}\natexlab{}.
\newblock \showarticletitle{Multi-turn response selection for chatbots with
  deep attention matching network}. In \bibinfo{booktitle}{\emph{Proceedings of
  the 56th Annual Meeting of the ACL (Volume 1: Long Papers)}}.
  \bibinfo{pages}{1118--1127}.
\newblock


\end{thebibliography}

\end{document}